\documentclass[11pt]{article}

\usepackage{acl}
\usepackage{todonotes}

\usepackage{latexsym}

\usepackage[export]{adjustbox}
\usepackage{fontawesome5}
\usepackage{microtype}
\usepackage{float}
\usepackage{listings}

\usepackage[english,bidi=default]{babel} % English as the main language.
\babelfont{rm}{TeXGyreTermesX} % similar to Times

\usepackage{booktabs}

\usepackage{graphicx}
\usepackage{subcaption}  % in preamble
\usepackage[dvipsnames,table]{xcolor}
\usepackage{xspace}
\usepackage{tcolorbox}
\usepackage{booktabs}
\usepackage{nicematrix}
\usepackage{pgfplots}
\pgfplotsset{compat=1.18}
\usepackage{fontspec}
\usepackage{tablefootnote} % for table footnotes
\newcommand{\rev}[1]{\textcolor{black}{#1}}
\newfontfamily\medievalfont{Junicode-Regular.ttf}[
    Path=./, 
    Scale=MatchLowercase
]
\newcommand{\ms}[1]{{\medievalfont #1}}

\title{When Simpler Is Better: \\Evaluating Translation Pipelines for Medieval Latin Manuscripts}

\author{Nguyen Kim Hai Bui$^{1,2}$\thanks{Equal contribution}, Md. Easin Arafat$^{2,3}$\footnotemark[1], Tam\'{a}s G\'{a}bor Orosz$^{2}$, {Mufti Mahmud}$^3$
\\
$^1$The Hong Kong Polytechnic University, $^2$E\"{o}tv\"{o}s Lor\'{a}nd University\\ $^3$King Fahd University of Petroleum and Minerals\\
\texttt{nguyen-kim-hai.bui@connect.polyu.hk\quad arafatmdeasin@inf.elte.hu}\\
 {\faIcon{github}}   \href{https://github.com/ELTE-DSED/Interpres}{\textbf{{ELTE-DSED/Interpres}}}\quad
 {\faIcon{database}}   \href{https://huggingface.co/collections/wnkh/interpres}{\textbf{{collections/wnkh/interpres}}}
}

\begin{document}
\maketitle

\begin{abstract}
Despite remarkable progress in machine translation, Vision Language Models (VLMs) struggle on historical manuscripts, a domain that stresses core Natural Language Processing (NLP) capabilities: low-resource transliteration, archaic vocabulary, and noisy input signals. We present a systematic framework for evaluating the full image-to-translation pipeline on medieval Latin manuscripts, a setting in which scribal shorthand, ligatures, and parchment degradation expose failure modes that are invisible in clean-text benchmarks. Benchmarking on the CATMuS Latin dataset reveals a specialization gap: domain-specific Optical Character Recognition (OCR) models reduce character error rate by up to 4.3$\times$ compared to general-purpose VLMs, despite operating at orders of magnitude fewer parameters. We introduce the Interpres-Parallel-Corpus (IPC), a novel dataset comprising 1,383 aligned manuscript image lines, transcriptions, and expert translations, \rev{to our knowledge} the first of its kind for medieval Latin. Our experiments uncover a complexity paradox: the simplest pipeline, a specialized OCR model feeding directly into a VLM, outperforms all multi-component variants. Adding retrieval-augmented generation (RAG) or post-OCR correction introduces prompt saturation and error propagation that degrade aggregate translation quality. These findings offer both a new benchmark and practical guidance for deploying translation systems in low-resource historical settings.
\end{abstract}

\section{Introduction}

\begin{figure}[!htbp]
    \centering
    \includegraphics[width=\linewidth]{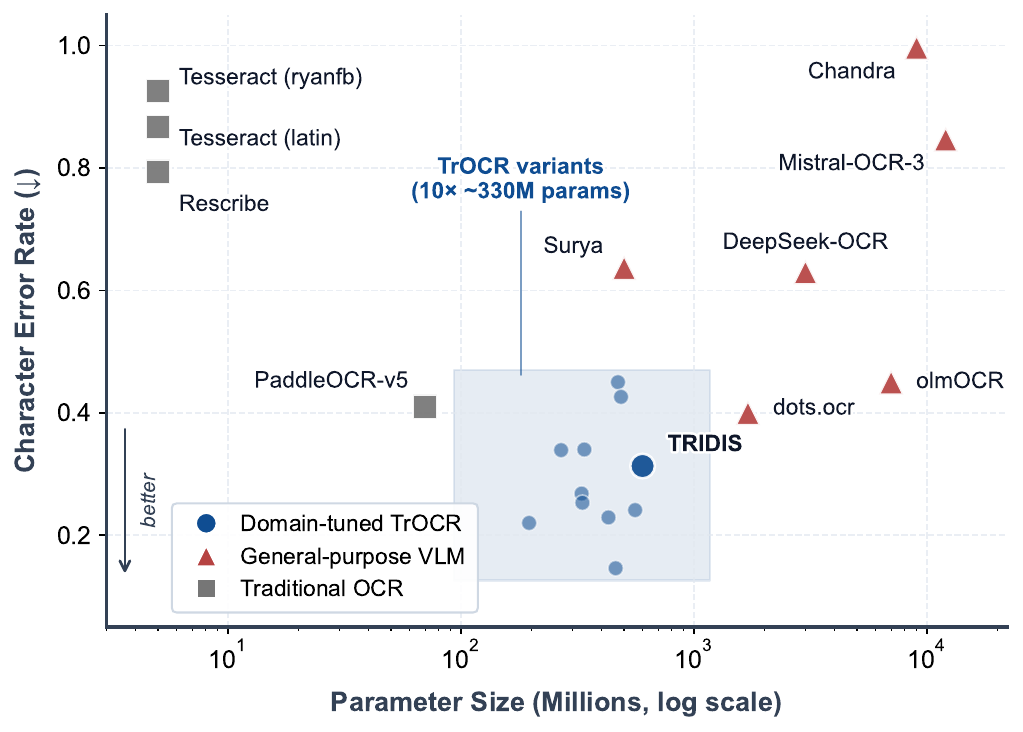}
    \caption{Scatter plot demonstrating the specialization gap. Small, domain-tuned TrOCR models achieve the lowest CER, outperforming general-purpose VLMs that are larger in size. Mistral-OCR-3 is assumed to be the largest, as the creator \rev{does not} publish its actual size.}
    \label{fig:ocr_scatter}
\end{figure}

Medieval Latin manuscripts are the foundational primary sources for European history, philosophy, and law from late antiquity to the early modern period \cite{Bischoff_1990}. From royal charters to philosophical treatises, these documents contain the blueprint of Western intellectual heritage. However, the vast majority of these collections remain untranscribed and untranslated, locked behind a paleographic barrier that requires years of specialized training to overcome. Historical text is not merely a humanities concern, it serves as a stress test for NLP robustness. Archaic orthographic conventions, low-resource scribal abbreviations, and morphological variation that deviates sharply from modern token distributions probe fundamental limitations in how current VLMs handle distributional shift. While some domains in machine translation have reached near-human parity \cite{hassan2018achieving}, historical manuscripts present a unique frontier defined by scribal abbreviations, non-standardized ligatures, and physical parchment degradation \cite{derolez2003palaeography}.

The typical approach to unlocking these texts involves a multi-stage process: OCR, post-OCR correction, and semantic translation. Recent advances in Transformer-based OCR (TrOCR) \cite{li_2023trocr} and character-level language modeling (ByT5) \cite{xue-etal-2022-byt5} have provided powerful building blocks for this task. However, there is a lack of systematic evaluation regarding how these components should be integrated into an end-to-end (E2E) pipeline. Specifically, it remains unclear whether massive VLMs can bypass the need for modularity, or whether more complex, retrieval-augmented architectures can handle the inherent noise of historical transcriptions.

In this paper, we introduce a modular framework designed to benchmark and optimize the journey from manuscript image to translated text. We address two primary research questions: (1) \textit{Does model specialization outperform massive parameter scale in historical paleography?} and (2) \textit{Do complex, multi-component pipelines consistently outperform simpler baselines on noisy historical data?} To answer these, we introduce the IPC, \rev{to our knowledge} the first tripartite dataset mapping medieval Latin manuscript images to both character-accurate transcriptions and authoritative English translations.

Our analysis reveals two striking phenomena that challenge modern AI scaling assumptions. First, we identify a specialization gap: domain-tuned models achieve a CER that is 4.3$\times$ lower than DeepSeek-OCR \cite{wei2025deepseekocrcontextsopticalcompression}, one of the largest dedicated OCR-focused VLMs evaluated, despite operating with orders-of-magnitude fewer parameters. This advantage is mechanistic rather than coincidental: TrOCR-Medieval-Base \cite{medieval_data_trocr_collection} is fine-tuned on CATMuS Medieval \cite{catmus}, a corpus of $\sim$195,000 annotated lines spanning nine centuries of Latin manuscripts, which provides direct exposure to paleographic variation, medieval scribal ligatures, and diachronic orthographic conventions that general-purpose VLMs lack. TRIDIS \cite{aguilar2025tridiscomprehensivemedievalearly} further specializes in documentary manuscripts (charters, registers, legal instruments) by training on semi-diplomatic transcription conventions, abbreviation expansion rules, and allograph normalization specific to these genres. The specialization gap thus reflects learned paleographic priors, not merely different architecture choices. Second, we identify a complexity paradox: a simple OCR-to-VLM baseline yields more reliable results than configurations that use RAG or post-OCR correction. We trace this failure to prompt saturation (i.e., high-density retrieval distracts the generator) and brittleness propagation (i.e., correction artifacts can interfere with downstream semantic reasoning). Our contributions are as follows:
\begin{itemize}
    \item We introduce the Interpres Parallel Corpus (IPC), a \rev{novel} benchmark of 1,383 aligned image lines, accurate transcriptions, and expert translations for medieval Latin research\rev{, to our knowledge the first to align all three modalities for this setting}.
    \item We explain the specialization gap mechanistically, showing that domain-exposure through paleographic corpora (CATMuS Medieval \cite{catmus}, TRIDIS \cite{aguilar2025tridiscomprehensivemedievalearly}) provides learned scribal priors that general-purpose VLMs cannot acquire through scale alone.
    \item We identify and quantify the complexity paradox, demonstrating that no multi-component variant consistently outperforms the simple OCR-to-VLM baseline, and we provide the first formal failure taxonomy for historical translation pipelines, characterizing two distinct failure mechanisms: prompt saturation and brittleness propagation. Unlike post-hoc pipeline comparisons, our taxonomy provides explanatory structure, predicting which pipeline configurations will degrade and why.
\end{itemize}

\section{Related Work}
\label{sec:related-work}

The digitization and automated analysis of historical manuscripts involve multiple converging frontiers in NLP and Computer Vision.

\paragraph{Historical Hand-written Text Recognition (HTR) and Corpora.} State-of-the-art HTR has transitioned from Convolutional Recurrent Neural Networks architectures utilizing Connectionist Temporal Classification \cite{graves2006ctc} to Transformer-based models like TrOCR \cite{li_2023trocr}\rev{, with tailored systems further improving recognition on medieval Latin specifically \cite{koch-etal-2023-tailored}}. This evolution has been supported by the emergence of large-scale, heterogeneous corpora. The TRIDIS corpus \cite{aguilar2025tridiscomprehensivemedievalearly} provides a unified resource of medieval and early modern documentary manuscripts, while CATMuS Medieval \cite{catmus} introduces a multilingual, diachronic dataset spanning nine centuries of Latin and vernacular scripts. These resources enable the development of the specialized OCR models analyzed in our specialization gap benchmark.

\paragraph{OCR Error Correction.} Post-OCR correction has evolved from rule-based and linguistic heuristics \cite{Springmann_2014} to neural approaches. Alfi and Way demonstrated that integrating neural correction modules can improve translation quality by up to 30\% \cite{afli-way-2016-integrating}. Modern strategies often leverage byte-level models like ByT5 \cite{xue-etal-2022-byt5} to handle the irregular orthography of historical texts, though the risk of error propagation in serialized pipelines remains high. In this work, we evaluate two ByT5 correction variants: $C_2$ (ByT5 yaya), which follows the modular pipeline of Momtaz et al. \cite{electronics14153083} using a fine-tuned ByT5 model for post-OCR correction on incunabula, and $C_1$ (ours), which adapts the $C_2$ architecture and training code to our setting by fine-tuning on OCR outputs from the specialization gap benchmark with CER ranging from 0.1 to 0.4. This CER range is to ensure the collected dataset doesn't capture too many gibberish cases from OCR models, and not too easily.

\paragraph{Historical and Low-Resource Machine Translation.} Machine Translation for dead or archaic languages presents unique data-scarcity challenges. Early efforts in Latin-to-English NMT reached BLEU scores of approximately 22.4 \cite{machinaCognoscens}. More recently, systems like LITERA \cite{rosu-2025-litera} and Hanja-to-Korean frameworks such as H2KE \cite{son-etal-2022-translating} have utilized large-scale pre-trained LLMs/VLMs with specialized fine-tuning to achieve expert-level performance. LITERA \cite{rosu-2025-litera}, the most directly related work, employs a fine-tuned GPT-4o-mini with multi-layer GPT-4o revision for Latin-to-English translation. However, LITERA assumes access to clean, pre-transcribed Latin text, a critical assumption that bypasses the core challenge of manuscript digitization: there is no OCR stage. In real-world manuscript workflows, transcribed text is unavailable; every document must first pass through character recognition before semantic translation can begin. Our framework evaluates the full image-to-translation cascade, revealing that OCR quality and pipeline topology critically determine translation outcomes for medieval manuscripts, findings that LITERA's clean-text evaluation design is structurally unable to capture. Crucially, we find that adding more components does not reliably improve results, framing the OCR integration problem as one of pipeline architecture rather than component replacement. Beyond Latin, low-resource historical NLP encompasses a range of under-resourced languages and scripts where the same specialization-versus-scale trade-off we investigate has been observed across diverse typological contexts \cite{10.1145/3764579,alshehhi2025inclusivenlpassessingcompressed,tukenov2026sozkztrainingefficientsmall}. Our work contributes an empirical data point to this broader question.

\paragraph{RAG Failure Modes and Context Distraction.} The integration of retrieval mechanisms into generation pipelines introduces well-documented failure modes. Our RAG system augments translation with historical lexical resources, including the Medieval Latin Word Vocabulary\footnote{
% \url{https://kriston.net/tools/latin/medieval.shtml}
% \url{https://anonymous.4open.science/r/medieval-latin-dicts-D540/medieval_latin_word_vocabulary.txt}
\url{https://huggingface.co/datasets/wnkh/medieval-latin-dicts/blob/main/mlw_vocabulary.txt}
}, the William Whitaker's Words\footnote{
% \url{https://anonymous.4open.science/r/medieval-latin-dicts-D540/WORDS.txt}
\url{https://huggingface.co/datasets/wnkh/medieval-latin-dicts/blob/main/WORDS.txt}
% \url{https://kriston.net/tools/latin/words.shtml}
}, the Lexicon Abbreviaturarum\footnote{\url{https://www.adfontes.uzh.ch/en/ressourcen/abkuerzungen/cappelli-online}} dictionary by Adriano Cappelli for paleographic shorthand, the Medieval Latin Dictionary\footnote{
% \url{http://dadako.narod.ru/knigohran/lat-eng_MedievalLatin_1_0.zip}
% \url{https://anonymous.4open.science/r/medieval-latin-dicts-D540/medieval_latin_dictionary.txt}
\url{https://huggingface.co/datasets/wnkh/medieval-latin-dicts/blob/main/mld.txt}
}, parallel Latin-English texts \cite{machinaCognoscens}, and the TRIDIS translation corpus \cite{aguilar2025tridiscomprehensivemedievalearly}. Retrieved passages are reranked by semantic similarity to the source OCR text before being appended to the prompt. Research into LLM/VLM context utilization reveals a lost-in-the-middle phenomenon, where models struggle to extract relevant information from dense, distractor-heavy contexts \cite{liu2023lostmiddlelanguagemodels}. In cross-lingual RAG applications, such as dictionary-augmented translation, providing excessive or tangentially relevant context can overwhelm the model's reasoning capabilities \cite{wu2024languagesequalinsightsmultilingual,li2025languagedriftmultilingualretrievalaugmented}. Specifically, dense retrieval (i.e., dense given context in the prompt) can trigger copy-back behavior, where the generator bypasses semantic translation and instead echoes the retrieved source tokens directly into the output \cite{zaranis-etal-2024-analyzing,ali-etal-2026-mitigating}.

\paragraph{Cascading Errors in Serialized NLP Pipelines.} Error propagation across pipeline stages has been identified as a structural challenge in multi-component NLP systems \cite{caselli_2015}. In machine translation pipelines, upstream noise, whether from ASR, OCR, or normalization modules, degrades downstream quality non-linearly when later components lack robustness to atypical input distributions \cite{todorov2022assessmentimpactocrnoise,shapira2025measuringeffecttranscriptionnoise,le-duc-etal-2025-multimed-st}. We extend this direction by quantifying both failure modes (prompt saturation and brittleness propagation) empirically in the context of medieval paleography, providing the first formal failure taxonomy for historical translation pipelines.

\section{The Specialization Gap}
\label{sec:specialization-gap}

\begin{figure}[!htbp]
    \centering
    \begin{tikzpicture}
        \begin{axis}[
            ybar,
            width=\linewidth,
            height=5cm,
            ylabel={CER $\downarrow$},
            symbolic x coords={Tesseract,Mistral,DeepSeek,Paddle,TRIDIS,TrOCR},
            xtick=data,
            xticklabel style={rotate=25,anchor=east},
            % nodes near coords,
            % nodes near coords align={vertical},
            grid=major,
            ymin=0, ymax=1
        ]
        \addplot+[fill=gray!30] coordinates {
            (Tesseract, 0.867) (Mistral, 0.847) (DeepSeek, 0.630) (Paddle, 0.410) (TRIDIS, 0.313) (TrOCR, 0.146)
        };
        \end{axis}
    \end{tikzpicture}
    \caption{The Specialization Gap. CER comparison across architectures on CATMuS Latin samples. Mistral: Mistral-OCR-3; \rev{DeepSeek: DeepSeek-OCR}; Paddle: PaddleOCR-v5; TrOCR: TrOCR-Medieval-Base. Domain-tuned models (TrOCR/TRIDIS) \rev{substantially} outperform massive VLMs despite being orders of magnitude smaller.}
    \label{fig:specialization_gap}
\end{figure}
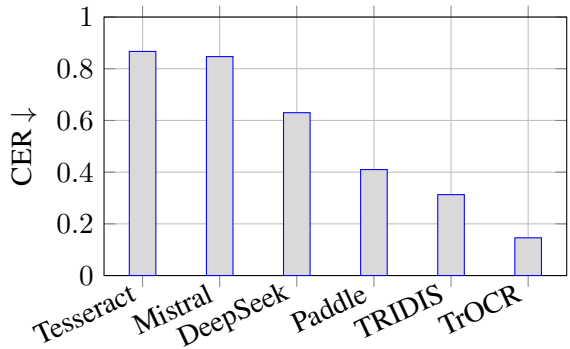

The initial stage of our pipeline extracts text from digitized manuscript images. Conventional AI scaling laws \cite{kaplan2020scalinglawsneurallanguage} suggest that increasing parameter counts, as seen in VLMs, should yield superior performance on edge cases like historical paleography. However, our benchmarks challenge this assumption, revealing a \rev{substantial} performance divergence we term the specialization gap.

\paragraph{Benchmarking Scale vs. Domain-Expertise.} We evaluated a range of OCR architectures against the Latin manuscripts from CATMuS dataset \cite{catmus} to identify the optimal vision foundation: traditional engines like Tesseract \cite{TessOverview,PageLayout,ScriptDetect,Multilingual,TableDetect}, Rescribe \cite{rescribe_ocr}, and PaddleOCR-v5 \cite{cui2025paddleocr30technicalreport}; massive general-purpose VLMs include DeepSeek-OCR \cite{wei2025deepseekocrcontextsopticalcompression}, Mistral-OCR-3 \cite{mistral_ocr3}, Surya \cite{paruchuri2025surya}, Chandra \cite{paruchuri2025chandra}, olmOCR-2 \cite{olmocr2}, dots.ocr \cite{li2025dotsocrmultilingualdocumentlayout}; and specialized historical models such as TrOCR Medieval models \cite{medieval_data_trocr_collection}, TRIDIS \cite{aguilar2025tridiscomprehensivemedievalearly}.

As shown in Figures~\ref{fig:ocr_scatter} and~\ref{fig:specialization_gap}, we observe that model size is an unreliable predictor of accuracy for medieval Latin. Massive general-purpose VLMs frequently fail to interpret scribal shorthand and ligatures. In contrast, specialized models, such as Base from the TrOCR Medieval collection, achieved a CER of 14.6\%, \rev{substantially} outperforming models with over 3B parameters. This demonstrates that for high-noise historical domains, specialized exposure is more valuable than sheer model scale. \rev{We revisit this ranking end-to-end in Appendix~\ref{sec:supp-analyses}: the lowest-CER engine is not the best translator, because CER against a diplomatic reference under-credits the abbreviation expansion that most aids the VLM (Table~\ref{tab:per-engine-cer}).}

\section{Evaluation Methodology}
\label{sec:methodology}

Quantifying E2E historical translation requires a codebase spanning vision, transcription, and semantics. We developed the framework to facilitate this systematic evaluation.

To support this study, we constructed the IPC. We sourced manuscript line images from the HTRomance project \cite{Clerice_HTRomance,glaise_2023_8362890}, specifically focusing on medieval Latin scripts from four literary works (Table~\ref{table:corpus_sources}). These were then aligned with expert-curated ground-truth transcriptions and translations from the Perseus Digital Library and Project Gutenberg. This tripartite dataset (Image Line $\leftrightarrow$ Medieval Latin $\leftrightarrow$ English) allows for granular error analysis at every stage of the pipeline.

\begin{table}[!htbp]
    \centering
    \setlength{\tabcolsep}{4pt}
    \begin{tabular}{llc}
        \toprule
        \textbf{Author} & \textbf{Work} & \textbf{Samples} \\
        \midrule
        \midrule
        Ovid & \textit{Metamorphoses}\tablefootnote{\url{https://catalog.perseus.org/catalog/urn:cts:latinLit:phi0959.phi006.perseus-eng1}} & 492 \\
        Catullus & \textit{Poems}\tablefootnote{\url{https://catalog.perseus.org/catalog/urn:cts:latinLit:phi0472.phi001.perseus-eng2}} & 166 \\
        Cicero & \textit{Tusculan Disputations}\tablefootnote{\url{https://www.gutenberg.org/ebooks/14988}} & 260 \\
        Quintilian & \textit{Institutio Oratoria}\tablefootnote{\url{https://catalog.perseus.org/catalog/urn:cts:latinLit:phi1002.phi001.perseus-eng2}} & 465 \\
        \midrule
        \textbf{Total} & & 1383 \\
        \bottomrule
    \end{tabular}
    \caption{Paleographic literary sources comprising the IPC. \textit{Metamorphoses}, \textit{Poems}, \textit{Institutio Oratoria} are taken from Perseus Digital Library and translated by B. More, L.C. Smithers, and H.E. Butler, respectively. \textit{Tusculan Disputations} is taken from Project Gutenberg and translated by C.D. Yonge.}
    \label{table:corpus_sources}
\end{table}

\paragraph{Dataset Construction and Annotation.}
\label{para:dataset}

\begin{figure*}[!htbp]
    \centering
    \usepgfplotslibrary{groupplots}
    \begin{tikzpicture}
        \begin{groupplot}[
            group style={
                group size=2 by 1,
                horizontal sep=1.5cm,
                xticklabels at=edge bottom
            },
            width=0.48\linewidth,
            height=5cm,
            grid=major,
            ymin=0, ymax=1.1
        ]
        \nextgroupplot[
            xlabel={Character Error Rate (CER)},
            ylabel={Density (Norm.)},
            legend style={at={(0.5,-0.4)},anchor=north,legend columns=4,font=\small},
            xmin=0, xmax=1.0
        ]
        % 1. AVG Full (bold blue solid) — length-weighted mu=0.451
        \addplot[blue, thick, solid, no marks] coordinates {
            (0.00,0.50) (0.05,0.58) (0.11,0.66) (0.16,0.75) (0.21,0.82) (0.26,0.89) (0.32,0.94) (0.37,0.98) (0.42,1.00) (0.47,1.00) (0.53,0.98) (0.58,0.95) (0.63,0.90) (0.68,0.83) (0.74,0.76) (0.79,0.68) (0.84,0.59) (0.89,0.51) (0.95,0.43) (1.00,0.36)
        }; \addlegendentry{Avg Full}

        % 7. PaddleOCR Full (violet, solid) — length-weighted mu=0.578
        \addplot[violet, thin, solid, no marks, opacity=0.45] coordinates {
            (0.00,0.05) (0.05,0.09) (0.11,0.14) (0.16,0.22) (0.21,0.31) (0.26,0.42) (0.32,0.55) (0.37,0.68) (0.42,0.81) (0.47,0.91) (0.53,0.98) (0.58,1.00) (0.63,0.98) (0.68,0.91) (0.74,0.80) (0.79,0.68) (0.84,0.55) (0.89,0.42) (0.95,0.31) (1.00,0.21)
        }; \addlegendentry{$O_1$ Full}

        % 5. TRIDIS Full (orange, solid) — length-weighted mu=0.424
        \addplot[orange, thin, solid, no marks, opacity=0.45] coordinates {
            (0.00,0.69) (0.05,0.75) (0.11,0.81) (0.16,0.86) (0.21,0.91) (0.26,0.95) (0.32,0.98) (0.37,0.99) (0.42,1.00) (0.47,0.99) (0.53,0.98) (0.58,0.95) (0.63,0.91) (0.68,0.87) (0.74,0.82) (0.79,0.76) (0.84,0.70) (0.89,0.63) (0.95,0.57) (1.00,0.50)
        }; \addlegendentry{$O_2$ Full}

        % 3. HTR Full (teal, solid) — length-weighted mu=0.351
        \addplot[teal, thin, solid, no marks, opacity=0.45] coordinates {
            (0.00,0.57) (0.05,0.67) (0.11,0.76) (0.16,0.85) (0.21,0.92) (0.26,0.97) (0.32,1.00) (0.37,1.00) (0.42,0.98) (0.47,0.94) (0.53,0.87) (0.58,0.79) (0.63,0.70) (0.68,0.60) (0.74,0.51) (0.79,0.42) (0.84,0.33) (0.89,0.26) (0.95,0.20) (1.00,0.15)
        }; \addlegendentry{$O_3$ Full}

        % 2. AVG Shared (bold red dashed) — length-weighted mu=0.413
        \addplot[red, thick, dashed, no marks] coordinates {
            (0.00,0.51) (0.05,0.60) (0.11,0.69) (0.16,0.77) (0.21,0.85) (0.26,0.92) (0.32,0.96) (0.37,0.99) (0.42,1.00) (0.47,0.99) (0.53,0.95) (0.58,0.90) (0.63,0.83) (0.68,0.75) (0.74,0.66) (0.79,0.57) (0.84,0.48) (0.89,0.40) (0.95,0.32) (1.00,0.26)
        }; \addlegendentry{Avg Shared}

        % 8. PaddleOCR Shared (magenta, dashed) — length-weighted mu=0.559
        \addplot[magenta!80!black, thin, dashed, no marks, opacity=0.45] coordinates {
            (0.00,0.07) (0.05,0.12) (0.11,0.18) (0.16,0.26) (0.21,0.36) (0.26,0.48) (0.32,0.61) (0.37,0.74) (0.42,0.86) (0.47,0.94) (0.53,0.99) (0.58,1.00) (0.63,0.96) (0.68,0.88) (0.74,0.77) (0.79,0.64) (0.84,0.51) (0.89,0.39) (0.95,0.28) (1.00,0.20)
        }; \addlegendentry{$O_1$ Shared}

        % 6. TRIDIS Shared (olive, dashed) — length-weighted mu=0.370
        \addplot[olive, thin, dashed, no marks, opacity=0.45] coordinates {
            (0.00,0.71) (0.05,0.78) (0.11,0.84) (0.16,0.89) (0.21,0.94) (0.26,0.97) (0.32,0.99) (0.37,1.00) (0.42,0.99) (0.47,0.97) (0.53,0.94) (0.58,0.90) (0.63,0.84) (0.68,0.78) (0.74,0.71) (0.79,0.64) (0.84,0.57) (0.89,0.50) (0.95,0.43) (1.00,0.37)
        }; \addlegendentry{$O_2$ Shared}

        % 4. HTR Shared (cyan, dashed) — length-weighted mu=0.311
        \addplot[cyan!70!black, thin, dashed, no marks, opacity=0.45] coordinates {
            (0.00,0.58) (0.05,0.69) (0.11,0.79) (0.16,0.88) (0.21,0.95) (0.26,0.99) (0.32,1.00) (0.37,0.98) (0.42,0.93) (0.47,0.86) (0.53,0.77) (0.58,0.67) (0.63,0.56) (0.68,0.46) (0.74,0.36) (0.79,0.28) (0.84,0.21) (0.89,0.15) (0.95,0.10) (1.00,0.07)
        }; \addlegendentry{$O_3$ Shared}

        \nextgroupplot[
            xlabel={Translation Quality (ChrF)},
            legend style={at={(0.5,-0.49)},anchor=north,legend columns=-1,font=\small},
            xmin=0, xmax=60
        ]
        % P0 Baseline (Mean=19.66, Std=14.13)
        \addplot+[ gray, opacity=0.3, mark=none, thick] coordinates {(0.0, 0.38) (2.5, 0.48) (5.0, 0.58) (7.5, 0.69) (10.0, 0.79) (12.5, 0.88) (15.0, 0.95) (17.5, 0.99) (20.0, 1.00) (22.5, 0.98) (25.0, 0.93) (27.5, 0.86) (30.0, 0.77) (32.5, 0.66) (35.0, 0.55) (37.5, 0.45) (40.0, 0.35) (42.5, 0.27) (45.0, 0.20) (47.5, 0.14) (50.0, 0.10) (52.5, 0.07) (55.0, 0.04) (57.5, 0.03) (60.0, 0.02)}; \addlegendentry{$P_0$}

        % P1 TRIDIS (Mean=26.45, Std=15.65)
        \addplot+[ blue, opacity=0.3, mark=none, thick] coordinates {(0.0, 0.24) (2.5, 0.31) (5.0, 0.39) (7.5, 0.48) (10.0, 0.58) (12.5, 0.67) (15.0, 0.77) (17.5, 0.85) (20.0, 0.92) (22.5, 0.97) (25.0, 1.00) (27.5, 1.00) (30.0, 0.98) (32.5, 0.93) (35.0, 0.86) (37.5, 0.78) (40.0, 0.69) (42.5, 0.59) (45.0, 0.50) (47.5, 0.41) (50.0, 0.32) (52.5, 0.25) (55.0, 0.19) (57.5, 0.14) (60.0, 0.10)}; \addlegendentry{$P_1$}

        % P2 TRIDIS+ByT5 (Mean=26.57, Std=14.92)
        \addplot+[ teal, opacity=0.3, mark=none, thick] coordinates {(0.0, 0.21) (2.5, 0.27) (5.0, 0.35) (7.5, 0.44) (10.0, 0.54) (12.5, 0.64) (15.0, 0.74) (17.5, 0.83) (20.0, 0.91) (22.5, 0.97) (25.0, 1.00) (27.5, 1.00) (30.0, 0.98) (32.5, 0.93) (35.0, 0.85) (37.5, 0.77) (40.0, 0.67) (42.5, 0.57) (45.0, 0.47) (47.5, 0.37) (50.0, 0.29) (52.5, 0.22) (55.0, 0.16) (57.5, 0.12) (60.0, 0.08)}; \addlegendentry{$P_2$}

        % P3 TRIDIS+RAG (Mean=26.11, Std=15.55)
        \addplot+[ orange, opacity=0.3, mark=none, thick] coordinates {(0.0, 0.24) (2.5, 0.32) (5.0, 0.40) (7.5, 0.49) (10.0, 0.59) (12.5, 0.68) (15.0, 0.78) (17.5, 0.86) (20.0, 0.93) (22.5, 0.98) (25.0, 1.00) (27.5, 1.00) (30.0, 0.97) (32.5, 0.92) (35.0, 0.85) (37.5, 0.77) (40.0, 0.67) (42.5, 0.58) (45.0, 0.48) (47.5, 0.39) (50.0, 0.31) (52.5, 0.24) (55.0, 0.18) (57.5, 0.13) (60.0, 0.09)}; \addlegendentry{$P_3$}

        % P4 TRIDIS+ByT5+RAG (Mean=26.24, Std=15.38)
        \addplot+[ Maroon, opacity=0.3, mark=none, thick] coordinates {(0.0, 0.23) (2.5, 0.30) (5.0, 0.39) (7.5, 0.48) (10.0, 0.57) (12.5, 0.67) (15.0, 0.77) (17.5, 0.85) (20.0, 0.92) (22.5, 0.97) (25.0, 1.00) (27.5, 1.00) (30.0, 0.97) (32.5, 0.92) (35.0, 0.85) (37.5, 0.77) (40.0, 0.67) (42.5, 0.57) (45.0, 0.48) (47.5, 0.39) (50.0, 0.30) (52.5, 0.23) (55.0, 0.17) (57.5, 0.13) (60.0, 0.09)}; \addlegendentry{$P_4$}
        \end{groupplot}
    \end{tikzpicture}
    \caption{(a) Fair Arena Representativeness Check. Length-weighted CER density for each OCR model over the full corpus and the Fair Arena subset. $O_1$ is Paddle, $O_2$ is TRIDIS, $O_3$ is TrOCR. (b) Performance Density Analysis. Individual pipeline distributions for topologies $P_0$ through $P_4$ on the Shared Benchmark, shown as per-sample sentence-level ChrF densities.}
    \label{fig:density_analysis}
\end{figure*}
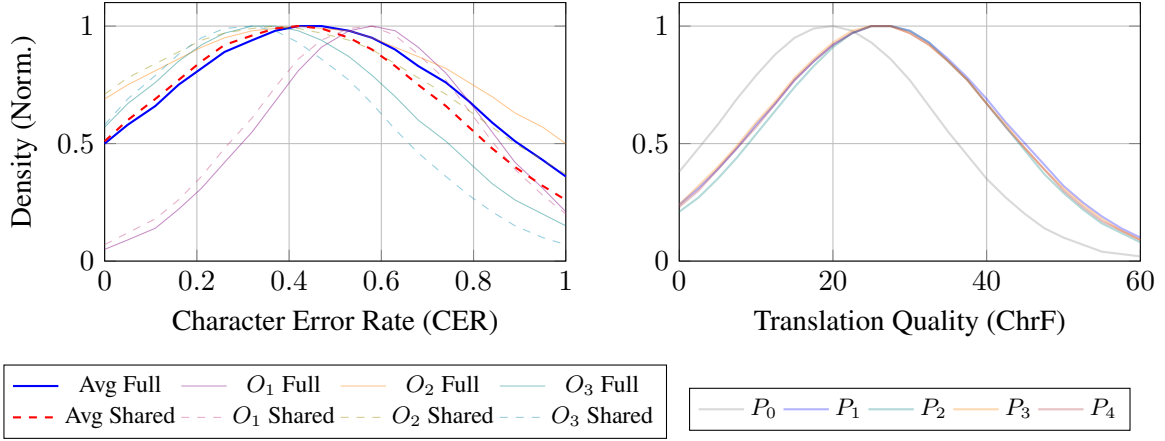

The IPC was constructed through a three-stage pipeline. First, manuscript line images and their corresponding OCR-text were sourced from the HTRomance project \cite{Clerice_HTRomance,glaise_2023_8362890}, which provides manually curated transcriptions for medieval Latin manuscripts. Second, authoritative English translations were identified at the chapter and page level from the Perseus Digital Library and Project Gutenberg, selecting editions whose translators are credited in Table~\ref{table:corpus_sources}. Third, using the given book/chapter/page numbers from HTRomance, individual manuscript lines were aligned with their corresponding translated sentence or clause in Gemini 3 Pro \cite{gemini_3}, which performed semantic matching between the Latin transcriptions and the target-language translation passages. Each alignment was constrained to a sliding window of the source text to prevent cross-chapter drift. We used a zero-shot prompt instructing the model to output only a single matched pair, and we verified alignment quality across all samples through manual inspection. The resulting triplets (image, Latin transcription, English translation) were retained for evaluation. \rev{Two properties bound the risk here: the model generates neither side of the pair and any residual misalignment is pipeline-independent, inflating variance uniformly rather than producing the differences we report.}

\begin{table}[!htbp]
    \centering
    \setlength{\tabcolsep}{4pt}
    \begin{tabular}{lrr}
        \toprule
        \textbf{Metric} & \textbf{Full Corpus} & \textbf{Fair Arena} \\
        \midrule
        \midrule
        Samples ($N$) & 1,383 & 1,003 \\
        Mean CER & 0.3507 & 0.3091 \\
        CER StdDev ($\sigma$) & 0.3556 & 0.3060 \\
        Mean Line Length & 38.6 & 39.6 \\
        Abbr. Density & 0.0912 & 0.0856 \\
        \bottomrule
    \end{tabular}
    \caption{Corpus statistics comparing the full IPC{} against the Fair Arena evaluation subset. Metrics confirm that the filtered subset maintains a representative difficulty profile across transcription noise (CER) and linguistic complexity (Length/Abbreviations).}
    \label{table:corpus_stats}
\end{table}

Despite its modest size of 1,383 samples (Table~\ref{table:corpus_stats}), the IPC is, to our knowledge, the first dataset to provide aligned image-transcription-translation triplets for medieval Latin. Prior datasets like CATMuS Medieval provide line-level transcriptions but no translations, whereas prior Latin translation work like LITERA uses only pre-transcribed text, with no manuscript images. The Fair Arena protocol (in the next paragraph) ensures rigorous comparison across pipeline topologies by restricting evaluation to the 1,003 samples successfully processed by all configurations, mitigating the confounding effects of unequal refusal rates across OCR engines. \rev{A full character-set inventory and the robustness of the specialization gap to CER normalization choices are reported in Appendix~\ref{sec:supp-analyses}.}

\paragraph{Fair Arena Benchmarking Protocol.} Evaluating commercial VLMs like GPT-4o \cite{openai2024gpt4ocard} on noisy historical data presents a unique challenge: adversarial refusal. We observed that extreme OCR noise (garbled character strings) often triggers automated safety guardrails. Specifically, models misinterpret these strings as potential adversarial injection attempts or trials to exploit Personally Identifiable Information due to the presence of historical names in the manuscripts. This leads to API Refusal Rates (RR) that vary across pipeline configurations, ranging from 1.3\% on clean text to nearly 24\% on aggressively noisy output.

To ensure statistical integrity, we implemented the fair arena protocol. We restricted our aggregate metrics to a shared benchmark subset of $N=1,003$ samples that were successfully processed across all candidate topologies. To verify the representativeness of this subset, we performed a Fair Arena Check by comparing the CER and length distributions of the shared subset against the full corpus ($N=1,383$), summarized in Table~\ref{table:corpus_stats}. CER is computed as the character-level edit distance between TrOCR-Medieval-Base predictions and the expert ground-truth transcriptions, normalized by reference length. We find that the shared subset maintains a comparable difficulty profile: the mean CER is $0.3091$ ($\sigma=0.306$) vs. $0.3507$ ($\sigma=0.355$) for the full corpus, and the mean line length is actually slightly higher in the shared subset ($39.6$ vs. $38.6$ characters). Furthermore, the density of medieval abbreviations (proxied by the fraction of non-alphanumeric, non-whitespace characters in the ground-truth transcription) remains nearly identical ($\mu=0.086$ vs. $0.091$). This confirms that the Fair Arena subset is a representative cross-section of the manuscript noise encountered in the full corpus, rather than an easier filtered subset. \rev{We further confirm that this exclusion does not favor simpler pipelines: penalizing refused outputs over the full corpus preserves, and widens, the $P_1$ advantage (Appendix~\ref{sec:supp-analyses}, Table~\ref{tab:refusal-penalized}).}

For all configurations, we use a structured system prompt that instructs the model to act as a medieval Latin paleography expert, cross-reference OCR and corrected text, and output a JSON object with the translation and optional notes. The RAG component retrieves from historical dictionaries and parallel texts using an embedding model and a cross-encoder reranker with an overfetch limit of 50 per source and returns the top 10 results. Full prompt text and retrieval configuration details are provided in Appendix~\ref{sec:interpres-prompt}.

Figure~\ref{fig:density_analysis}(a) visualizes this representativeness check as a set of length-weighted CER kernel density estimates. For each OCR architecture (TrOCR-Medieval-Base, TRIDIS, PaddleOCR), we compute the character error rate of every prediction against the expert ground-truth Latin transcription from our corpus, then fit a Gaussian with mean $\mu_w = \frac{\sum_i \ell_i \cdot \text{CER}_i}{\sum_i \ell_i}$ and variance weighted by line length $\ell_i = |\text{ref}_i|$. Weighting by length ensures that each character contributes equally to the distribution, so short lines with high noise do not disproportionately skew the estimate. The two bold curves show the pooled averages across all three architectures for the full corpus (solid blue, $\mu_w=0.451$) and the Fair Arena subset (dashed red, $\mu_w=0.413$); their close agreement across the entire CER range confirms that our filtering criterion preserves the full difficulty spectrum.

\section{E2E Pipeline Results}
\label{sec:results}

We evaluated five pipeline topologies, each progressively augmenting the input to GPT-4o (using default OpenAI API sampling parameters): $P_0$ receives only the manuscript image; $P_1$ receives the image and the raw OCR text; $P_2$ receives the image, OCR text, ByT5-corrected text; $P_3$ receives the image, OCR text, reranked retrieval context from historical dictionaries and parallel texts; and $P_4$ combines all preceding components. For clarity, $P_1$ corresponds to $O_2$, $P_2$ to $O_2$+$C_1$, $P_3$ to $O_2$+$R$, and $P_4$ to $O_2$+$C_1$+$R$ in Tables~\ref{tab:pipeline-comprehensive} and~\ref{tab:fair-arena-leaderboard}. To maintain fairness across commercial model refusal behaviors, all main results are reported on the Fair Arena shared subset ($N=1,003$). The performance differences between $P_1$ and $P_2$ (paired t-test, $t=4.76$, $p<0.001$) and between $P_1$ and $P_4$ ($t=3.71$, $p<0.001$) are statistically significant; the difference between $P_1$ and $P_3$ is not significant ($t=1.33$, $p=0.18$), indicating that RAG alone does not reliably improve translation quality. \rev{These COMET-level differences correspond to small effect sizes (Cohen's $d_z \le 0.15$), and on ChrF the differences are negligible with $95\%$ bootstrap confidence intervals that all include zero (Appendix~\ref{sec:supp-analyses}, Table~\ref{tab:effect-sizes}).} Figure~\ref{fig:density_analysis}(b) shows the per-sample ChrF distributions for each topology.

\begin{table}[h]
    \centering
    \small
    \setlength{\tabcolsep}{5.5pt}
    \begin{tabular}{ccccc}
        \toprule
        \textbf{Pipeline} & \textbf{RR}$^{\%\downarrow}$  & $\textbf{COMET}^\uparrow$ & $\textbf{BLEU}^\uparrow$ & $\textbf{ChrF}^\uparrow$ \\
        \midrule
        \midrule
         Baseline & 1.30 & 0.4537 & 3.44 & 18.81 \\
         \midrule
         $O_1$  & 17.86 & 0.4651 & 3.61 & 19.36 \\
         $O_2$  & \color{Maroon}\textbf{2.53} & \color{Maroon}\textbf{0.5100} & \color{Maroon}\textbf{5.76} & \color{Maroon}\textbf{24.93} \\
         $O_3$  & 2.53 & 0.5003 & 5.22 & 22.81 \\
         \midrule
         $O_1$ + $C_1$  & 16.99 & 0.4504 & 3.44 & 18.62 \\
         $O_1$ + $C_2$  & 23.93 & 0.4515 & 3.50 & 18.65 \\
         $O_2$ + $C_1$  & \color{Maroon}2.75 &\color{Maroon} 0.5033 &\color{Maroon} 5.17 &\color{Maroon} 24.87  \\
         $O_2$ + $C_2$  & 4.19 & 0.5021 & 5.23 & 24.39  \\
         $O_3$ + $C_1$  & 4.77 & 0.4931 & 5.02 & 22.32  \\
         $O_3$ + $C_2$  & 4.84 & 0.4892 & 4.44 & 22.17  \\
         \midrule
         $O_1$ + $R$  & 12.80 & 0.4531 & 3.35 & 18.77  \\
         $O_2$ + $R$  & \color{Maroon}3.25 &\color{Maroon} 0.5073 &\color{Maroon} 5.15 &\color{Maroon} 24.17  \\
         $O_3$ + $R$  & 3.69 & 0.4931 & 4.73 & 21.89  \\
         \midrule
         $O_1$ + $C_1$ + $R$  & -- & -- & -- & -- \\
         $O_1$ + $C_2$ + $R$  & -- & -- & -- & -- \\
         $O_2$ + $C_1$ + $R$  & 4.56 &\color{Maroon} 0.5045 & 5.18 &\color{Maroon} 24.56 \\
         $O_2$ + $C_2$ + $R$  & \color{Maroon}3.98 & 0.5012 & \color{Maroon}5.20 & 24.36  \\
         $O_3$ + $C_1$ + $R$  & 5.71 & 0.4897 & 4.64 & 22.21  \\
         $O_3$ + $C_2$ + $R$  & 3.76 & 0.4883 & 4.70 & 22.06  \\
        \bottomrule
    \end{tabular}
    \caption{Comprehensive evaluation results across all individual execution runs for each pipeline topology. ($O_1$: PaddleOCR, $O_2$: TRIDIS, $O_3$: TrOCR Medieval Base, $C_1$: ByT5 (ours), $C_2$: ByT5 (yaya), $R$: RAG. The $\uparrow$ symbol indicates that higher scores are better, while $\downarrow$ indicates that lower scores are better. RR stands for refusal rate, the percentage of samples where the API refused to return a response due to safety guardrails. The full table is in the Appendix~\ref{sec:appendix}.}
    \label{tab:pipeline-comprehensive}
\end{table}

\begin{table}[!htbp]
\centering
\small
\setlength{\tabcolsep}{5.5pt}
\begin{tabular}{lccc}
\toprule
\textbf{Pipeline} & \textbf{BLEU$^\uparrow$} & \textbf{ChrF$^\uparrow$} & \textbf{COMET$^\uparrow$} \\
\midrule
\midrule
 Baseline ($P_0$) & 3.73 & 19.68 & 0.4615 \\
 OCR ($P_1$) & \color{Maroon}\textbf{6.26} & \color{Maroon}\textbf{26.00} & \color{Maroon}\textbf{0.5195} \\
 + Correct ($P_2$) & 5.67 & 25.83 & 0.5108 \\
 + RAG ($P_3$) & 5.75 & 25.38 & 0.5169 \\
 + Correct + RAG ($P_4$) & 5.84 & 25.58 & 0.5122 \\
 \bottomrule
\end{tabular}
\caption{Fair Arena Leaderboard: E2E pipeline performance comparison (N=1,003). All pipelines in this table (except baseline $P_0$) utilize the TRIDIS engine as the primary vision engine. The correction variant used in this table is $C_1$.}
\label{tab:fair-arena}
\end{table}

\begin{figure}[h]
    \centering
    \includegraphics[width=\linewidth]{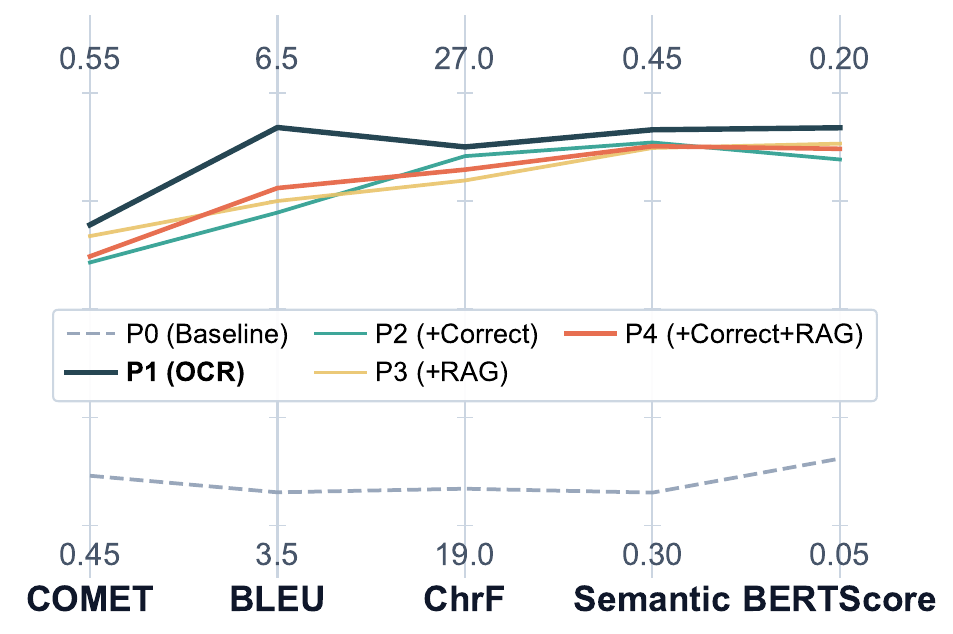}
    \caption{Parallel coordinates chart comparing the top pipeline topologies across five normalized NLP metrics. While $P_1$ dominates exact-match metrics, the full pipeline $P_4$ remains highly competitive on semantic metrics.}
    \label{fig:pipeline_radar}
\end{figure}

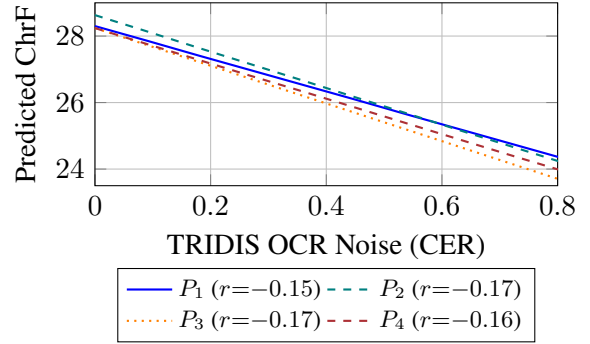
\begin{figure}[!htbp]
    \centering
    \begin{tikzpicture}
        \begin{axis}[
            width=\linewidth, height=4cm,
            xlabel={TRIDIS OCR Noise (CER)},
            ylabel={Predicted ChrF},
            legend style={at={(0.5,-0.45)},anchor=north,legend columns=2,font=\small},
            grid=major, xmin=0, xmax=0.8, ymin=23.5, ymax=29
        ]
        % P1 O2+GPT4o: corr=-0.146 slope=-4.92 b=28.30
        \addplot[blue, thick, solid] coordinates {
            (0.00,28.30) (0.09,27.86) (0.18,27.41) (0.27,26.97) (0.36,26.53) (0.45,26.09) (0.54,25.64) (0.63,25.20) (0.72,24.76) (0.81,24.32)
        }; \addlegendentry{$P_1$ ($r{=}{-}0.15$)}
        % P2 O2+C1+GPT4o: corr=-0.172 slope=-5.49 b=28.63
        \addplot[teal, thick, dashed] coordinates {
            (0.00,28.63) (0.09,28.14) (0.18,27.64) (0.27,27.15) (0.36,26.66) (0.45,26.16) (0.54,25.67) (0.63,25.17) (0.72,24.68) (0.81,24.19)
        }; \addlegendentry{$P_2$ ($r{=}{-}0.17$)}
        % P3 O2+R+GPT4o: corr=-0.170 slope=-5.67 b=28.24
        \addplot[orange, thick, dotted] coordinates {
            (0.00,28.24) (0.09,27.73) (0.18,27.22) (0.27,26.71) (0.36,26.20) (0.45,25.69) (0.54,25.18) (0.63,24.67) (0.72,24.16) (0.81,23.65)
        }; \addlegendentry{$P_3$ ($r{=}{-}0.17$)}
        % P4 O2+C1+R+GPT4o: corr=-0.161 slope=-5.31 b=28.24
        \addplot[Maroon, thick, dashed] coordinates {
            (0.00,28.24) (0.09,27.76) (0.18,27.28) (0.27,26.80) (0.36,26.33) (0.45,25.85) (0.54,25.37) (0.63,24.89) (0.72,24.41) (0.81,23.94)
        }; \addlegendentry{$P_4$ ($r{=}{-}0.16$)}
        \end{axis}
    \end{tikzpicture}
    \caption{Noise robustness of all TRIDIS-based pipeline topologies. Lines show sentence-level ChrF predicted by linear regression over TRIDIS CER. All four configurations exhibit nearly identical degradation slopes ($\approx{-5}$ ChrF/CER, $r \approx -0.15$), confirming that pipeline depth does not affect noise sensitivity.}
    \label{fig:complexity_paradox}
\end{figure}

\begin{figure}[h]
    \centering
    \includegraphics[width=\linewidth]{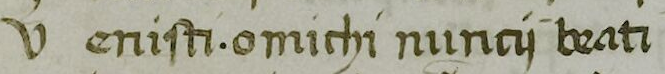}
    \caption{Archetype A (Synergy). Latin: \ms{Uenisti. o michi nuncii beati}. Translation: \textit{You have come back. O joyful news to me!}}
    \label{fig:eval-synergy}
    \vspace{0.5em}
    \centering\small
    \setlength{\tabcolsep}{3pt}
    \begin{tabular}{cl}
        & \multicolumn{1}{c}{\textbf{Translation Output}} \\
        \midrule
        \midrule
        $P_0$ & {O Christ, now bless.} \\
        $P_1$ & {And you, sent forth, the announcement of the blessed.} \\
        $P_2$ & {You have come, blessed messenger.} \\
        $P_3$ & \textbf{You have come. O joyful news to me!} \\
        $P_4$ & \textbf{{You have come, O joyful news to me.}} \\
        \bottomrule
    \end{tabular}
    \captionof{table}{Performance delta for Figure~\ref{fig:eval-synergy}. $P_3$ and $P_4$ resolve the idiomatic ``news'' through RAG retrieval.}
    \label{table:eval-synergy}
\end{figure}

\paragraph{The Complexity Paradox.} Our results (Table~\ref{tab:fair-arena} and Figure~\ref{fig:pipeline_radar}) reveal a counter-intuitive finding we term the Complexity Paradox: the simplest specialized pipeline achieves the highest mean ChrF of 26.00, and no multi-component variant reliably improves upon it. Adding correction ($P_2$: 25.83), RAG ($P_3$: 25.38), or both ($P_4$: 25.58) yields no statistically reliable gain over the simple baseline. Notably, $P_4$ remains competitive on semantic-level metrics (Figure~\ref{fig:pipeline_radar}), and a small subset of samples show RAG providing genuine lexical disambiguation benefits (Figure~\ref{fig:eval-synergy}, Table~\ref{table:eval-synergy}), but these cases do not outweigh the aggregate complexity overhead. As shown in Figure~\ref{fig:complexity_paradox}, all four TRIDIS-based topologies exhibit virtually identical noise sensitivity against OCR noise (corr $\approx -0.15$ to $-0.17$, slope $\approx -5$ ChrF/CER), confirming that the VLM absorbs transcription errors equally regardless of pipeline depth. The observed variance between $P_1$ and $P_2$ to $P_4$ is therefore attributable to two component-specific failure modes detailed below, not to differential noise sensitivity. \rev{Consistent with this, a gold-transcription oracle improves ChrF by only ${\sim}2$ points over $P_1$ (Appendix~\ref{sec:supp-analyses}, Table~\ref{tab:oracle}), leaving little OCR-attributable headroom for correction or retrieval to recover.}

\begin{figure}[h]
    \centering
    \includegraphics[width=\linewidth]{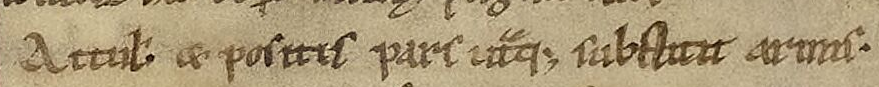}
    \caption{Archetype B (Paradox). Latin: \ms{Attuł. \& positis pars itͣq substitit armis.} Translation: \textit{while both sides resting, laid aside their arms.}}
    \label{fig:eval-paradox}
    \vspace{0.5em}
    \centering\small
    \setlength{\tabcolsep}{3pt}
    \begin{tabular}{cp{0.9\linewidth}}
        & \multicolumn{1}{c}{\textbf{Translation Output}} \\
        \midrule
        \midrule
        $P_0$ & {And he, as a powerful protector of peace, subdued the arms.} \\
        $P_1$ & \textbf{{And with the arms laid aside, each side paused.}} \\
        $P_2$ & {Having been brought and with the positions set, each side halted.} \\
        $P_3$ & {Attul et positis pars utraque substitit.} \\
        $P_4$ & {The troop, having been set in position, stood on each side.} \\
        \bottomrule
    \end{tabular}
    \captionof{table}{The RAG Paradox in Figure~\ref{fig:eval-paradox}. Dense retrieval triggers an ``echo failure'' in $P_3$.}
    \label{table:eval-paradox}
\end{figure}

\paragraph{Prompt Saturation.} In $P_3$, the inclusion of high-density dictionary retrieval often acts as a cognitive distractor. Instead of synthesizing a translation based on context, the model exhibits a copy-back behavior, echoing the Latin source text or dictionary definitions directly into the English output (Figure~\ref{fig:eval-paradox}, Table~\ref{table:eval-paradox}). Quantitative analysis reveals a weak positive correlation between RAG token overlap and translation quality ($r=0.146$, explaining only $\sim$2\% of variance on the TRIDIS-based $P_3$ run), suggesting that while RAG provides useful lexical anchors in some cases, the model can overly rely on direct extraction from the retrieval context at the expense of syntactic coherence.

\begin{figure}[h]
    \centering
    \includegraphics[width=\linewidth]{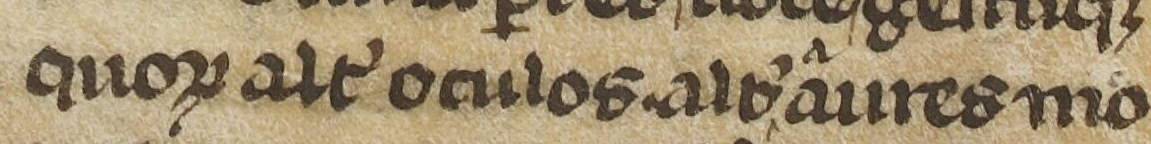}
    \caption{Archetype C (Cascade). Latin: \ms{quoꝵ alt̾ oculos. alt̾ͣ aures mo}. Translation: \textit{of which the one appeals to the eye and the other to the ear.}}
    \label{fig:eval-cascade}
    \vspace{0.5em}
    \centering\small
    \setlength{\tabcolsep}{3pt}
    \begin{tabular}{cp{0.9\linewidth}}
        & \multicolumn{1}{c}{\textbf{Translation Output}} \\
        \midrule
        \midrule
        $P_0$ & {how now, O Paulus, the white ears} \\
        $P_1$ & \textbf{{from which one moved the eyes, the other the ears.}} \\
        $P_2$ & {parts \texttimes{}22 \ldots{} from whom one has eyes, another has ears} \\
        $P_3$ & {parts \texttimes{}12 \ldots{} from which one has eyes the other ears} \\
        $P_4$ & {parts \texttimes{}22 \ldots{} of which one has eyes the other ears} \\
        \bottomrule
    \end{tabular}
    \captionof{table}{Cascade failure in Figure~\ref{fig:eval-cascade} caused by ByT5 repetition loops.}
    \label{table:eval-cascade}
\end{figure}

\paragraph{Brittleness Propagation.} In $P_2$ and $P_4$, character-level repetition loops introduced by the ByT5 correction layer can interfere with downstream translation. When these loops occur, the VLM is forced to reason over corrupted strings, leading to degraded translation quality (Figure~\ref{fig:eval-cascade}, Table~\ref{table:eval-cascade}). We detect repetition artifacts in $0.60\%$ of samples in the TRIDIS-based $P_2$ configuration, where they incur an average penalty of $7.15$ ChrF points compared to non-repeating samples ($\mu_{\text{rep}}=19.47$ vs.\ $\mu_{\text{non-rep}}=26.61$). Notably, this failure mode accounts for only a small fraction of the overall performance gap between $P_1$ and $P_2$, suggesting that additional degradation mechanisms (e.g., minor hallucinations introduced by imperfect correction) also contribute to the complexity paradox.

\paragraph{Diagnostic Archetypes.} To make these dynamics concrete, we organize the qualitative evidence into three diagnostic archetypes: \textit{Synergy} (Figure~\ref{fig:eval-synergy}), where RAG resolves lexical or idiomatic ambiguity that the simpler pipelines miss; \textit{Paradox} (Figure~\ref{fig:eval-paradox}), where dense retrieval triggers the prompt-saturation echo failure described above; and \textit{Cascade} (Figure~\ref{fig:eval-cascade}), where correction-layer artifacts propagate into the VLM and corrupt the final translation. Extended examples for each archetype are provided in Appendix~\ref{sec:archetype-examples}.

\section{Conclusion}
\label{sec:conclusion}

This paper shows that for medieval manuscript translation, model specialization and pipeline simplicity are paramount. We introduce the IPC and identify the specialization gap and the complexity paradox. Our findings suggest that current generic VLMs are not yet capable of replacing specialized historical OCR systems, and that complex multi-component pipelines may suffer from error propagation. Future work will explore whether dynamic pipeline selection based on OCR confidence can recover the occasional benefits of retrieval without incurring the aggregate complexity penalty.
\rev{Pipeline simplicity also carries a practical advantage independent of backend: Adding a correction model, embedding model, reranker, and retrieval datastore inflating prompt text and per-call cost. Since the quality benefit is at best small, $P_1$ remains the pragmatic default for deployment.}

\section*{Limitations}
First, the dataset is restricted to Medieval Latin, which does not capture the full variance of paleographic challenges in other historical scripts. \rev{Moreover, IPC is built from four classically transmitted literary works and therefore under-represents documentary and non-literary medieval Latin such as charters and registers; our claims concern the image-to-translation processing of Latin manuscripts rather than medieval Latin language modeling in general.} 

Second, our RAG relies on static dictionary anchors; future work could explore dynamic retrieval from larger contextual corpora to resolve deeper ambiguities. \rev{A systematic ablation over retrieval depth, sources, context volume, and correction strategies is left to future work.} 

Third, our main pipeline evaluations use GPT-4o as the translation backend. \rev{Preliminary experiments with additional open-source VLM backends suggest that the relative contribution of the correction and retrieval components may be backend-dependent: retrieval alone shows no significant benefit over the simple pipeline across the backends we tested, but the net effect of combining correction and retrieval appears to vary with a backend's capacity to arbitrate multiple noisy input signals. A systematic study of this backend-dependence, controlling for prompt format, decoding strategy, and model capacity, is an important direction for future work.} 

\rev{Finally, our gold-transcription oracle shows that perfect transcription raises end-to-end quality by only ${\sim}2$ ChrF over the best OCR pipeline, indicating that the dominant bottleneck is not OCR noise but the translation stage itself, together with the looseness of the available literary reference translations. Progress in historical manuscript translation therefore depends as much on richer, more literal parallel resources and on evaluation better suited to literary translation as on further OCR or vision-model gains. Our evaluation is also fully automatic: since all topologies share the same references, looseness acts as noise rather than a bias, so relative comparisons are more robust than absolute scores. A human evaluation with multiple annotators and adjudication, and a dedicated alignment audit reporting agreement and error rates, remain important future work.}

\section*{Ethics Statement}
DH research involving historical manuscripts carries an ethical responsibility to preserve cultural heritage with high fidelity. While our pipeline improves translation accessibility, we caution that model hallucinations in the context of fragmented or rare texts can lead to significant historical misinterpretations. Our tools are designed to augment, not replace, expert human inquiry. We commit to open-sourcing the IPC to foster transparent and reproducible benchmarking in this domain.

\bibliography{custom}

\appendix

\section{Usage of AI Assistants}
We used a large language model for editorial tasks such as grammar correction and enhancing clarity and readability. We also use a language model to align the translations as discussed in \S\ref{sec:methodology}.

\section{Detailed Configurations}
\label{sec:interpres-prompt}

\subsection{Translation System Prompt}
\label{sec:system-prompt}

The following system prompt is used for all GPT-4o-based translation configurations. The prompt is designed to leverage expertise in medieval Latin paleography while handling noisy OCR input and retrieval-augmented context.

\begin{lstlisting}[language=Python, frame=single, basicstyle=\small\ttfamily, breaklines=true]
You are Interpres, an expert in medieval Latin paleography and translation.

You will receive:
- **OCR Text (raw)** (when available): Text extracted directly from a manuscript image via OCR. This may contain character-level errors.
- **Suggested Corrected Text** (when available): The same text after automated correction by a ByT5 model. This may improve accuracy but can also introduce new errors, especially if the text is too long or the OCR already contains errors.
- **Dictionary Matches** (when available): Relevant entries from Latin dictionaries (high reliability).
- **Parallel Text Matches** (when available): Similar passages from Latin-English parallel corpora (reference only).

Your task:
1. **Cross-reference** the raw OCR and suggested corrected versions. Where they disagree, use context, grammar, and the reference materials to determine the most likely original text.
2. **Translate** the reconstructed Latin text faithfully.
3. If there are ambiguous or damaged sections, note them briefly.

You MUST respond with a JSON object containing exactly two keys:
- "translation": The English translation of the Latin text.
- "note": Brief notes about ambiguous readings, damaged sections, or translation choices. Use an empty string if there are no notes.
\end{lstlisting}

\subsection{Translation Configuration}
\label{sec:translation-config}

We use GPT-4o (model: \texttt{gpt-4o-2024-08-06}) with the default OpenAI API sampling parameters; the response format is JSON, and image detail is high. All pipelines are executed via the BatchAPI, with one pipeline per batch. The total API cost across all experiments is about \$50.

\subsection{RAG Configuration}
\label{sec:rag-config}

\begin{table}[h]
\centering
\small
\setlength{\tabcolsep}{3pt}
\begin{tabular}{lc}
    \toprule
    \textbf{Parameter} & \textbf{Value} \\
    \midrule
    \midrule
    Overfetch Limit (per source) & 50 \\
    Final Top-K after Reranking & 10 \\
    Embedding Batch Size & 64 \\
    Retrieval Sources & Dictionary, Parallel Texts \\
    Embedding Model & all-MiniLM-L6-v2\tablefootnote{\url{https://huggingface.co/sentence-transformers/all-MiniLM-L6-v2}} \\
    Reranking Model & ms-marco-MiniLM-L-6-v2\tablefootnote{\url{https://huggingface.co/cross-encoder/ms-marco-MiniLM-L6-v2}} \\
    \bottomrule
\end{tabular}
\caption{RAG retrieval and reranking configuration.}
\label{tab:rag-config}
\end{table}

The RAG component augments the prompt with relevant dictionary entries and parallel text passages. Dictionaries include Medieval Latin Word Vocabulary\tablefootnote{
% \url{https://kriston.net/tools/latin/medieval.shtml}
% \url{https://anonymous.4open.science/r/medieval-latin-dicts-D540/medieval_latin_word_vocabulary.txt}
\url{https://huggingface.co/datasets/wnkh/medieval-latin-dicts/blob/main/mlw_vocabulary.txt}
}, the William Whitaker's Words\tablefootnote{
% \url{https://kriston.net/tools/latin/words.shtml}
% \url{https://anonymous.4open.science/r/medieval-latin-dicts-D540/WORDS.txt}
\url{https://huggingface.co/datasets/wnkh/medieval-latin-dicts/blob/main/WORDS.txt}
}, the Lexicon Abbreviaturarum\tablefootnote{\url{https://www.adfontes.uzh.ch/en/ressourcen/abkuerzungen/cappelli-online}}, the Medieval Latin Dictionary\tablefootnote{
% \url{http://dadako.narod.ru/knigohran/lat-eng_MedievalLatin_1_0.zip}
% \url{https://anonymous.4open.science/r/medieval-latin-dicts-D540/medieval_latin_dictionary.txt}
\url{https://huggingface.co/datasets/wnkh/medieval-latin-dicts/blob/main/mld.txt}
}. Parallel Texts include Latin-English texts \cite{machinaCognoscens} and the TRIDIS translation corpus \cite{aguilar2025tridiscomprehensivemedievalearly}. The pipeline overfetches 50 candidates per source, deduplicates by text content, reranks using a cross-encoder, and returns the top 10 results for prompt augmentation. Key parameters are summarized in Table~\ref{tab:rag-config}.

\subsection{ByT5 Correction Model Training}
\label{sec:byt5-training}

The $C_1$ correction model is a fine-tuned ByT5-Small (\texttt{google/byt5-small}) that maps noisy OCR outputs to corrected Latin transcriptions. Training was performed on a single NVIDIA L40S GPU (CUDA 12.8) using the Transformers Trainer API. Key hyperparameters are listed in Table~\ref{tab:byt5-config}. The model was trained for 7 epochs with early stopping based on validation CER, saving up to 3 checkpoints. The model will be publicly released on Hugging Face upon acceptance.

\begin{table}[!htbp]
\centering
\small
\begin{tabular}{lc}
    \toprule
    \textbf{Parameter} & \textbf{Value} \\
    \midrule
    \midrule
    Base Model & ByT5-Small \\
    Effective Batch Size & $4\times4\times1$ \\
    Learning Rate & $3 \times 10^{-4}$ \\
    Epochs & 7 \\
    Hardware & 1 $\times$ NVIDIA L40S \\
    Training Framework & Transformers 4.57.6 \\
    \bottomrule
\end{tabular}
\caption{OCR correction fine-tuning configuration.}
\label{tab:byt5-config}
\end{table}

\section{Detailed Experimental Results}
\label{sec:appendix}

Table~\ref{tab:full-pipeline-comprehensive} reports the full pipeline evaluation across all OCR engines and correction strategies. Table~\ref{tab:fair-arena-leaderboard} provides the extended Fair Arena leaderboard with additional metrics.

\begin{table*}[!htbp]
    \centering
    \small
    \begin{tabular}{ccccccc}
        \toprule
        \textbf{Config.} & \textbf{RR}$^{\%\downarrow}$  & $\textbf{COMET}^\uparrow$ & $\textbf{BLEU}^\uparrow$ & $\textbf{ChrF}^\uparrow$ & $\textbf{Semantic}^\uparrow$ & $\textbf{BERTScore}^\uparrow$ \\
        \midrule
        \midrule
         Baseline & 1.30 & 0.4537 & 3.44 & 18.81 & 0.2912 & 0.0608 \\
         \midrule
         $O_1$  & 17.86 & 0.4651 & 3.61 & 19.36 & 0.3179 & 0.0861 \\
         $O_2$  & \color{Maroon}\textbf{2.53} & \color{Maroon}\textbf{0.5100} & \color{Maroon}\textbf{5.76} & \color{Maroon}\textbf{24.93} & \color{Maroon}\textbf{0.4140} & \color{Maroon}\textbf{0.1727} \\
         $O_3$  & \color{Maroon}{2.53} & 0.5003 & 5.22 & 22.81 & 0.3907 & 0.1540 \\
         \midrule
         $O_1$ + $C_1$  & 16.99 & 0.4504 & 3.44 & 18.62 & 0.3009 & 0.0802 \\
         $O_1$ + $C_2$  & 23.93 & 0.4515 & 3.50 & 18.65 & 0.3030 & 0.0693 \\
         $O_2$ + $C_1$  & \color{Maroon}2.75 & \color{Maroon}0.5033 & \color{Maroon}5.17 & \color{Maroon}24.87 & \color{Maroon}0.4107 & \color{Maroon}0.1657 \\
         $O_2$ + $C_2$  & 4.19 & 0.5021 & 5.23 & 24.39 & 0.4084 & 0.1628 \\
         $O_3$ + $C_1$  & 4.77 & 0.4931 & 5.02 & 22.32 & 0.3877 & 0.1476 \\
         $O_3$ + $C_2$  & 4.84 & 0.4892 & 4.44 & 22.17 & 0.3795 & 0.1434 \\
         \midrule
         $O_1$ + $R$  & 12.80 & 0.4531 & 3.35 & 18.77 & 0.3015 & 0.0723 \\
         $O_2$ + $R$  & \color{Maroon}3.25 & \color{Maroon}0.5073 & \color{Maroon}5.15 & \color{Maroon}24.17 & \color{Maroon}0.4083 & \color{Maroon}0.1689 \\
         $O_3$ + $R$  & 3.69 & 0.4931 & 4.73 & 21.89 & 0.3812 & 0.1435 \\
         \midrule
         $O_1$ + $C_1$ + $R$  & -- & -- & -- & -- & -- & -- \\
         $O_1$ + $C_2$ + $R$  & -- & -- & -- & -- & -- & -- \\
         $O_2$ + $C_1$ + $R$  & 4.56 & \color{Maroon}0.5045 & 5.18 & \color{Maroon}24.56 & \color{Maroon}0.4082 & \color{Maroon}0.1701 \\
         $O_2$ + $C_2$ + $R$  & 3.98 & 0.5012 & \color{Maroon}5.20 & 24.36 & 0.4022 & 0.1638 \\
         $O_3$ + $C_1$ + $R$  & 5.71 & 0.4897 & 4.64 & 22.21 & 0.3802 & 0.1415 \\
         $O_3$ + $C_2$ + $R$  & \color{Maroon}3.76 & 0.4883 & 4.70 & 22.06 & 0.3759 & 0.1381 \\
        \bottomrule
    \end{tabular}
    \caption{Comprehensive evaluation results across all individual execution runs for each pipeline topology. ($O_1$: PaddleOCR, $O_2$: TRIDIS, $O_3$: TrOCR Medieval Base, $C_1$: ByT5 (ours) fine-tuned on specialization-gap OCR outputs (CER 0.1-0.4), $C_2$: ByT5 (yaya) from \cite{electronics14153083}, $R$: RAG). The $\uparrow$ symbol indicates that higher scores are better, while $\downarrow$ indicates that lower is better. BERTScore is rescaled. RR stands for refusal rate, the percentage of samples where the API refused to return a response due to safety guardrails.}
    \label{tab:full-pipeline-comprehensive}
\end{table*}

\begin{table*}
    \centering
    \begin{tabular}{cccccc}
        \toprule
        \textbf{Config.} & $\textbf{COMET}^\uparrow$ & $\textbf{BLEU}^\uparrow$ & $\textbf{ChrF}^\uparrow$ & $\textbf{Semantic}^\uparrow$ & $\textbf{BERTScore}^\uparrow$ \\
        \midrule
        \midrule
         Baseline & 0.4615 & 3.73 & 19.68 & 0.3114 & 0.0732 \\
         \midrule
         $O_2$  & \color{Maroon}{\textbf{0.5195}} & \color{Maroon}{\textbf{6.26}} & \color{Maroon}{\textbf{26.00}} & \color{Maroon}{\textbf{0.4372}} & \color{Maroon}{\textbf{0.1879}} \\
         $O_3$  & 0.5074 & 5.85 & 23.94 & 0.4136 & 0.1635 \\
         \midrule
         $O_2$ + $C_1$  & \color{Maroon}{0.5108} & 5.67 & \color{Maroon}{25.83} & \color{Maroon}{0.4328} & \color{Maroon}{0.1769} \\
         $O_2$ + $C_2$  & 0.5094 & \color{Maroon}{5.68} & 25.35 & 0.4289 & 0.1742 \\
         $O_3$ + $C_1$  & 0.4995 & 5.49 & 23.26 & 0.4076 & 0.1579 \\
         $O_3$ + $C_2$  & 0.4954 & 4.85 & 22.99 & 0.3996 & 0.1557 \\
         \midrule
         $O_2$ + $R$  & \color{Maroon}{0.5169} & \color{Maroon}{5.75} & \color{Maroon}{25.38} & \color{Maroon}{0.4309} & \color{Maroon}{0.1824} \\
         $O_3$ + $R$  & 0.5017 & 5.26 & 23.03 & 0.4035 & 0.1579 \\
         \midrule
         $O_2$ + $C_1$ + $R$  & \color{Maroon}{0.5122} & \color{Maroon}{5.84} & \color{Maroon}{25.58} & \color{Maroon}{0.4315} & \color{Maroon}{0.1806} \\
         $O_2$ + $C_2$ + $R$  & 0.5105 & 5.69 & 25.54 & 0.4243 & 0.1787 \\
         $O_3$ + $C_1$ + $R$  & 0.4980 & 5.03 & 23.23 & 0.4028 & 0.1545 \\
         $O_3$ + $C_2$ + $R$  & 0.4972 & 5.26 & 23.16 & 0.3998 & 0.1539 \\
        \bottomrule
    \end{tabular}
    \caption{Fair Arena. ($O_2$: TRIDIS, $O_3$: TrOCR Medieval Base, $C_1$: ByT5 (ours) fine-tuned on specialization-gap OCR outputs, $C_2$: ByT5 (yaya) from \cite{electronics14153083}, $R$: RAG). The $\uparrow$ symbol indicates that higher scores are better. BERTScore is rescaled. The bold-red line shows the best pipeline configuration in the arena, red lines show the best pipelines in their own group of topology.}
    \label{tab:fair-arena-leaderboard}
\end{table*}

\section{Supplementary Analyses}
\label{sec:supp-analyses}

\paragraph{Per-engine IPC transcription vs.\ downstream translation.} Table~\ref{tab:per-engine-cer} reports, for each OCR engine, character/word error rates against the diplomatic gold transcription, the character composition of the engine output, and the resulting raw-OCR translation quality. TrOCR-Medieval attains the lowest CER \emph{and} WER because it faithfully reproduces the diplomatic abbreviations of the reference (non-ASCII character fraction $0.078$, close to the gold $0.085$). TRIDIS instead \emph{expands} scribal abbreviations into plain ASCII Latin (non-ASCII fraction $0.001$), which increases its measured CER against the diplomatic reference yet supplies the most translation-ready input to the VLM and the best downstream ChrF. Word-level accuracy mirrors CER, so this is not a tokenization artifact. This reconciles the apparent paradox that the lowest-CER engine is not the best translator: CER against a diplomatic reference under-credits normalization that is useful downstream, so transcription-stage CER is a necessary but not sufficient predictor of end-to-end quality.

\begin{table}[!htbp]
\centering\small
\setlength{\tabcolsep}{3pt}
\begin{tabular}{lccccc}
\toprule
\textbf{Engine} & \textbf{CER}$\downarrow$ & \textbf{WER}$\downarrow$ & \textbf{Abbr.} & \textbf{n-ASCII} & \textbf{ChrF}$\uparrow$ \\
\midrule\midrule
Gold (ref.) & -- & -- & 0.091 & 0.085 & -- \\
\midrule
$O_1$ Paddle & 0.586 & 0.955 & 0.022 & 0.026 & 19.36 \\
$O_3$ TrOCR  & \textbf{0.351} & \textbf{0.727} & 0.085 & 0.078 & 22.81 \\
$O_2$ TRIDIS & 0.447 & 0.937 & 0.024 & 0.001 & \textbf{24.93} \\
\bottomrule
\end{tabular}
\caption{Per-engine IPC transcription quality (full corpus) vs.\ downstream translation. CER/WER are against the diplomatic gold; Abbr.\ is the non-alphanumeric character fraction of the engine output; n-ASCII is its non-ASCII fraction; ChrF is the raw-OCR pipeline. TrOCR minimizes CER/WER by reproducing diplomatic abbreviations, yet abbreviation-expanding TRIDIS (n-ASCII $0.001$) translates best.}
\label{tab:per-engine-cer}
\end{table}

\paragraph{Gold-transcription oracle (OCR vs.\ translation error).} To separate transcription error from translation error, we translate the \emph{gold} Latin transcriptions directly with the same GPT-4o backend. Table~\ref{tab:oracle} reports the result on the Fair Arena. Perfect transcription raises ChrF by only ${\sim}1.9$ points over the noisy-OCR pipeline $P_1$ ($26.00\!\rightarrow\!27.87$), bounding the OCR-attributable share of end-to-end error as small and consistent with the noise-robustness slope in Figure~\ref{fig:complexity_paradox} (${\approx}{-}5$ ChrF/CER $\times$ mean CER $0.31 \approx 1.5$ ChrF). Two consequences follow. First, even with perfect input the absolute ceiling is low (ChrF $27.87$, BLEU $7.62$), confirming that the modest scores throughout this paper reflect the looseness of literary reference translations rather than pipeline failure. Second, a backend that had memorized the published translations would echo them when handed gold Latin and inflate BLEU into the tens; the small oracle BLEU instead indicates genuine translation rather than recall. The (image-free) ``ceiling'' and ``$P_1$ framing'' rows differ only in whether the gold text is labelled as trustworthy or as ``raw OCR'', isolating a small ${\sim}0.9$ ChrF framing effect.

\begin{table}[!htbp]
\centering\small
\setlength{\tabcolsep}{6pt}
\begin{tabular}{lcc}
\toprule
\textbf{Condition} & \textbf{ChrF}$^\uparrow$ & \textbf{BLEU}$^\uparrow$ \\
\midrule\midrule
$P_1$ (noisy OCR) & 26.00 & 6.26 \\
Oracle (gold, $P_1$ framing) & 26.99 & 6.92 \\
Oracle (gold, ceiling) & \textbf{27.87} & \textbf{7.62} \\
\bottomrule
\end{tabular}
\caption{Gold-transcription oracle on the Fair Arena ($N{=}1{,}003$). Translating the expert gold Latin with the same GPT-4o backend lifts ChrF by only ${\sim}2$ points over $P_1$, bounding the OCR-attributable error; the low absolute ceiling reflects loose literary references, and the small BLEU argues against training-data memorization of the published translations.}
\label{tab:oracle}
\end{table}

\paragraph{Memorization / contamination probe.} Because the reference translations come from public editions (Perseus, Project Gutenberg), we test whether GPT-4o reproduces them from memory rather than translating. Using a recall-oriented prompt that explicitly asks for the exact published translation of each passage (citing its author and work), we measure verbatim overlap with the reference across all $1{,}383$ samples. Reproduction is negligible: only $0.2\%$ of reference $8$-grams and $0.3\%$ of full lines are reproduced verbatim, and the longest common contiguous word run averages $10\%$ of the reference length, even under explicit recall prompting. The normal (oracle) translation shows the same negligible overlap ($0.1\%$ of $8$-grams). The published translations are therefore not retrievable from the model, so contamination does not inflate our results; the modest oracle scores (Table~\ref{tab:oracle}) reflect genuine translation of a difficult literary register, not recall.

\paragraph{Effect sizes and confidence intervals.} Table~\ref{tab:effect-sizes} complements the significance tests in \S\ref{sec:results} with paired effect sizes (Cohen's $d_z$) and $95\%$ bootstrap confidence intervals ($10^4$ resamples) for the difference between $P_1$ and each augmented pipeline on the Fair Arena. On sentence-level ChrF every interval includes zero and $|d_z|\le0.04$: the added components produce no practically meaningful change. On COMET the $P_1$ advantage over $P_2$ and $P_4$ is statistically reliable but the effect size remains small ($d_z\le0.15$), while $P_3$ is indistinguishable from $P_1$. Together these quantify the Complexity Paradox: augmentation does not help, and where it significantly differs it slightly \emph{hurts}.

\begin{table}[h]
\centering\small
\setlength{\tabcolsep}{4pt}
\begin{tabular}{llccc}
\toprule
\textbf{Metric} & \textbf{Pair} & $\Delta$ & \textbf{95\% CI} & $d_z$ \\
\midrule\midrule
ChrF  & $P_1{-}P_2$ & $-0.12$ & $[-0.65,+0.40]$ & $-0.01$ \\
ChrF  & $P_1{-}P_3$ & $+0.34$ & $[-0.19,+0.88]$ & $+0.04$ \\
ChrF  & $P_1{-}P_4$ & $+0.21$ & $[-0.35,+0.75]$ & $+0.02$ \\
\midrule
COMET & $P_1{-}P_2$ & $+0.009$ & $[+0.005,+0.012]$ & $+0.15$ \\
COMET & $P_1{-}P_3$ & $+0.003$ & $[-0.001,+0.006]$ & $+0.04$ \\
COMET & $P_1{-}P_4$ & $+0.007$ & $[+0.003,+0.011]$ & $+0.12$ \\
\bottomrule
\end{tabular}
\caption{Paired effect sizes (Cohen's $d_z$) and $95\%$ bootstrap CIs ($10^4$ resamples) for $P_1$ minus each augmented pipeline on the Fair Arena ($N{=}1{,}003$); ChrF is sentence-level. ChrF differences are negligible and all CIs include zero; COMET differences for $P_2$/$P_4$ are reliable but small.}
\label{tab:effect-sizes}
\end{table}

\paragraph{Character set and CER normalization.} The gold transcriptions use $103$ unique characters ($45$ non-ASCII), spanning combining diacritics (e.g., U+0303 \textsc{combining tilde}, $1{,}810$ occurrences), MUFI Private-Use codepoints (e.g., U+F1AC, $454$ occurrences), and dedicated abbreviation glyphs (e.g., U+A770 \textsc{modifier letter us}). Transcriptions are stored in diplomatic form and are intentionally not NFC-folded, preserving combining marks for paleographic fidelity. Table~\ref{tab:cer-norm} shows that the specialization ranking (TrOCR $<$ TRIDIS $<$ Paddle) is invariant to CER normalization (raw; NFC + casefold; NFC + casefold + punctuation/whitespace stripping), with CER shifting by at most $0.03$; the specialization gap is therefore not an artifact of normalization choices.

\begin{table}[h]
\centering\small
\setlength{\tabcolsep}{5pt}
\begin{tabular}{lccc}
\toprule
\textbf{Engine} & \textbf{Raw} & \textbf{NFC+lc} & \textbf{+strip} \\
\midrule\midrule
$O_3$ TrOCR  & \textbf{0.351} & \textbf{0.354} & \textbf{0.336} \\
$O_2$ TRIDIS & 0.447 & 0.475 & 0.462 \\
$O_1$ Paddle & 0.586 & 0.584 & 0.573 \\
\bottomrule
\end{tabular}
\caption{IPC CER under three normalizations: raw; NFC $+$ casefold (lc); and additionally stripping punctuation and collapsing whitespace (+strip). The engine ranking is invariant and CER shifts by $\le0.03$.}
\label{tab:cer-norm}
\end{table}

\paragraph{Accounting for refusals.} The Fair Arena restricts evaluation to samples processed by all pipelines. To verify that this does not favor simpler pipelines, Table~\ref{tab:refusal-penalized} reports full-corpus ($N{=}1{,}383$) ChrF where refused or invalid outputs are scored as empty, together with per-pipeline coverage. The simplest pipeline $P_1$ remains best, and its margin over $P_3$/$P_4$ \emph{widens} relative to the Fair Arena, because the more complex pipelines also refuse more often (coverage $95.4$--$97.5\%$). The Complexity Paradox thus holds, indeed strengthens, once refusals are penalized rather than excluded.

\begin{table}[h]
\centering\small
\setlength{\tabcolsep}{5pt}
\begin{tabular}{lccc}
\toprule
\textbf{Pipeline} & \textbf{Cov.\%} & \textbf{ChrF$_{\text{fair}}$} & \textbf{ChrF$_{\text{pen}}$} \\
\midrule\midrule
$P_0$ Baseline & 98.7 & 19.68 & 18.68 \\
$P_1$ OCR      & 97.5 & \textbf{26.00} & \textbf{24.65} \\
$P_2$ +Corr    & 97.3 & 25.83 & 24.56 \\
$P_3$ +RAG     & 96.7 & 25.38 & 23.69 \\
$P_4$ +Both    & 95.4 & 25.58 & 23.73 \\
\bottomrule
\end{tabular}
\caption{Refusal-penalized robustness. ChrF$_{\text{fair}}$ is the Fair Arena ($N{=}1{,}003$); ChrF$_{\text{pen}}$ scores refused/invalid outputs as empty over the full corpus ($N{=}1{,}383$). $P_1$ stays best and its margin over $P_3$/$P_4$ widens, as complex pipelines refuse more (lower coverage).}
\label{tab:refusal-penalized}
\end{table}

\color{black}

\section{Detailed Qualitative Case Studies}
\label{sec:archetype-examples}

This appendix provides extended qualitative evidence for the three diagnostic archetypes discussed in Section~\ref{sec:results}. For each archetype, we show 5 representative examples with the original Latin source, expert reference translation, and pipeline outputs.

{
\begin{table*}
    \small\centering
    \setlength{\tabcolsep}{3pt}\begin{tabular}{m{0.1\linewidth}p{0.8\linewidth}}
        \textbf{523} & \textbf{Content} \\ \midrule\midrule
        Image & \includegraphics[width=\linewidth, valign=c]{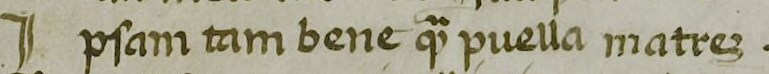} \\
        Latin & \ms{Ipsam tam bene ꝙͣ puella matre.} \\
        English &  as well as a girl knows her own mother. \\ 
        Archetype & Synergy \\ \midrule
        $P_0$ (Base) & psalm as well as the girl mother \\
        $P_1$ (OCR) & [She] herself as well as the girl, [as] mother. \\
        $P_2$ (Corr) & herself as well as the girl mothers. \\
        $P_3$ (RAG) & \textbf{As well as a girl knows her mother.} \\
        $P_4$ (Full) & \textbf{She knew her as well as a girl knows her own mother.} \\
    \end{tabular}
\end{table*}

\begin{table*}
    \small\centering
    \setlength{\tabcolsep}{3pt}\begin{tabular}{m{0.1\linewidth}p{0.8\linewidth}}
        \textbf{524} & \textbf{Content} \\ \midrule\midrule
        Image & \includegraphics[width=\linewidth, valign=c]{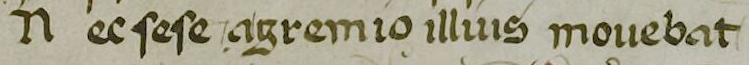} \\
        Latin & \ms{Nec sese a gremio illius mouebat} \\
        English &  Nor did it move from her lap, \\ 
        Archetype & Synergy \\ \midrule
        $P_0$ (Base) & He did not move from her embrace. \\
        $P_1$ (OCR) & Nor did he/she move from his/her embrace. \\
        $P_2$ (Corr) & nor did he/she move from the embrace of that person. \\
        $P_3$ (RAG) & \textbf{Nor did he move from her embrace.} \\
        $P_4$ (Full) & \textbf{Nor would he move from her lap.} \\
    \end{tabular}
\end{table*}

\begin{table*}
    \small\centering
    \setlength{\tabcolsep}{3pt}\begin{tabular}{m{0.1\linewidth}p{0.8\linewidth}}
        \textbf{648} & \textbf{Content} \\ \midrule\midrule
        Image & \includegraphics[width=\linewidth, valign=c]{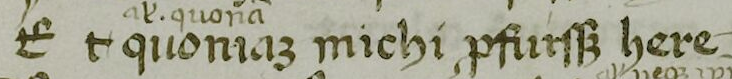} \\
        Latin & \ms{Et quonia michi ꝓfuiss here} \\
        English &  and whether I had made any money there. \\ 
        Archetype & Synergy \\ \midrule
        $P_0$ (Base) & And because it pleased me to begin here \\
        $P_1$ (OCR) & And since to me it first appeared \\
        $P_2$ (Corr) & And because to me it seemed here \\
        $P_3$ (RAG) & \textbf{And because it is necessary for me to bear here.} \\
        $P_4$ (Full) & \textbf{And because it was beneficial for me here.} \\
    \end{tabular}
\end{table*}

\begin{table*}
    \small\centering
    \setlength{\tabcolsep}{3pt}\begin{tabular}{m{0.1\linewidth}p{0.8\linewidth}}
        \textbf{274} & \textbf{Content} \\ \midrule\midrule
        Image & \includegraphics[width=\linewidth, valign=c]{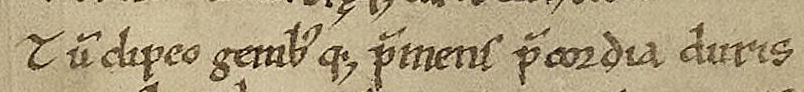} \\
        Latin & \ms{Tũ clipeo genibꝰ q p̾mens p̾ cordia duris} \\
        English &  Then pressing with buckler and hard knees the breast of Cygnus, \\ 
        Archetype & Synergy \\ \midrule
        $P_0$ (Base) & You, with a deceitful heart, always pondering cruel things. \\
        $P_1$ (OCR) & Et cum dixero, genibusque primens, prima cordia duris. \\
        $P_2$ (Corr) & And when I shall speak, pressing my knees, the first hearts are hard. \\
        $P_3$ (RAG) & \textbf{And you, pressing your knees, endure the hard hearts first.} \\
        $P_4$ (Full) & \textbf{And when I say, pressing first with my knees, the first hearts are hardened.} \\
    \end{tabular}
\end{table*}

\begin{table*}
    \small\centering
    \setlength{\tabcolsep}{3pt}\begin{tabular}{m{0.1\linewidth}p{0.8\linewidth}}
        \textbf{901} & \textbf{Content} \\ \midrule\midrule
        Image & \includegraphics[width=\linewidth, valign=c]{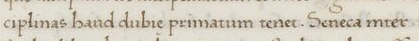} \\
        Latin & \ms{ciplinas haud dubie primatum tenet. Seneca inter} \\
        English &  disciplines hold the primacy without doubt. Seneca, among \\ 
        Archetype & Synergy \\ \midrule
        $P_0$ (Base) & Certainly, he undoubtedly held the first place. Seneca says: \\
        $P_1$ (OCR) & Without a doubt, [he/she/it] holds the first place in disciplines. Seneca among [them/us]. \\
        $P_2$ (Corr) & It undoubtedly holds the first place among the disciplines. Among them is Seneca. \\
        $P_3$ (RAG) & \textbf{Disciplines undoubtedly hold the first place. Seneca among them.} \\
        $P_4$ (Full) & \textbf{Disciplines undoubtedly hold the first place. Seneca among them.} \\
    \end{tabular}
\end{table*}

}

{
\begin{table*}
    \small\centering
    \setlength{\tabcolsep}{3pt}\begin{tabular}{m{0.1\linewidth}p{0.8\linewidth}}
        \textbf{1252} & \textbf{Content} \\ \midrule\midrule
        Image & \includegraphics[width=\linewidth, valign=c]{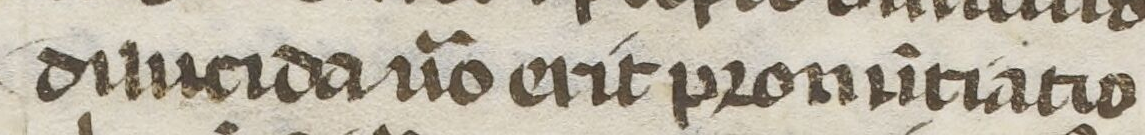} \\
        Latin & \ms{dilucida u̾o erit pronũtiatio} \\
        English &  The delivery will be clear if, \\ 
        Archetype & Paradox \\ \midrule
        $P_0$ (Base) & dilucidatio art [proprietates] \\
        $P_1$ (OCR) & \textbf{the explanation will indeed be clear} \\
        $P_2$ (Corr) & Indeed, the pronunciation will be clear. \\
        $P_3$ (RAG) & The elucidation, however, will be a pronouncement. \\
        $P_4$ (Full) & Indeed, the pronunciation will be clear. \\
    \end{tabular}
\end{table*}

\begin{table*}
    \small\centering
    \setlength{\tabcolsep}{3pt}\begin{tabular}{m{0.1\linewidth}p{0.8\linewidth}}
        \textbf{120} & \textbf{Content} \\ \midrule\midrule
        Image & \includegraphics[width=\linewidth, valign=c]{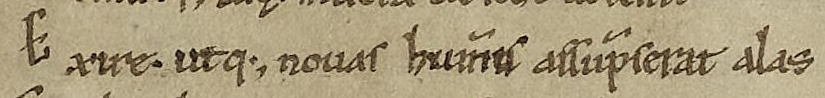} \\
        Latin & \ms{Exire. utq nouas hum̾is assũpserat alas} \\
        English &  “As he assumed upon his shoulders wings newformed, \\ 
        Archetype & Paradox \\ \midrule
        $P_0$ (Base) & From either side, now wings lift this one up. \\
        $P_1$ (OCR) & \textbf{And it was as if new wings had taken on the human condition.} \\
        $P_2$ (Corr) & 1473. To go forth and take up new wings of the earth... \\
        $P_3$ (RAG) & E xire utque novas huminis assumpserat alas \\
        $P_4$ (Full) & 1473. He began to take on new wings of humanity. \\
    \end{tabular}
\end{table*}

\begin{table*}
    \small\centering
    \setlength{\tabcolsep}{3pt}\begin{tabular}{m{0.1\linewidth}p{0.8\linewidth}}
        \textbf{636} & \textbf{Content} \\ \midrule\midrule
        Image & \includegraphics[width=\linewidth, valign=c]{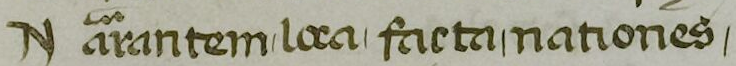} \\
        Latin & \ms{Naͬrantem / loca / facta / nationes /} \\
        English &  of regions, deeds, and peoples... \\ 
        Archetype & Paradox \\ \midrule
        $P_0$ (Base) & Not all places are blessed. \\
        $P_1$ (OCR) & \textbf{Narrating places, deeds, nations.} \\
        $P_2$ (Corr) & Telling places, deeds, nations \\
        $P_3$ (RAG) & Telling stories, places made, nations \\
        $P_4$ (Full) & Telling the places, deeds, nations \\
    \end{tabular}
\end{table*}

\begin{table*}
    \small\centering
    \setlength{\tabcolsep}{3pt}\begin{tabular}{m{0.1\linewidth}p{0.8\linewidth}}
        \textbf{357} & \textbf{Content} \\ \midrule\midrule
        Image & \includegraphics[width=\linewidth, valign=c]{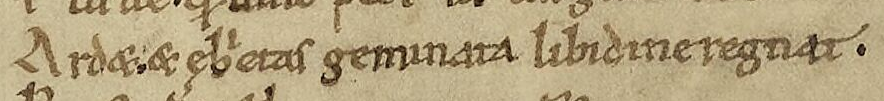} \\
        Latin & \ms{Ard\&. \& ębͥetas geminata libidine regnat.} \\
        English &  inflamed with drunkenness and lust \\ 
        Archetype & Paradox \\ \midrule
        $P_0$ (Base) & Ardisce, do not allow twin desires to rule over you. \\
        $P_1$ (OCR) & \textbf{Greed and drunkenness reign, doubled with lust.} \\
        $P_2$ (Corr) & Greed and drunkenness, twinned by lust, reign. \\
        $P_3$ (RAG) & Greed and laziness, doubled by lust, reign. \\
        $P_4$ (Full) & Arrogance and drunkenness, joined together, rule by lust. \\
    \end{tabular}
\end{table*}

\begin{table*}[H]
    \small\centering
    \setlength{\tabcolsep}{3pt}\begin{tabular}{m{0.1\linewidth}p{0.8\linewidth}}
        \textbf{1064} & \textbf{Content} \\ \midrule\midrule
        Image & \includegraphics[width=\linewidth, valign=c]{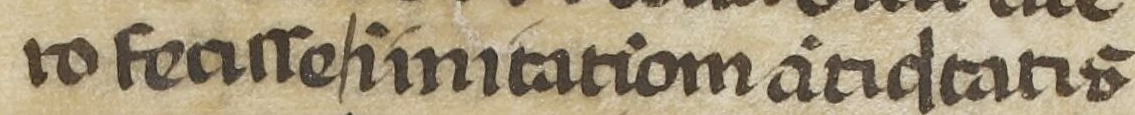} \\
        Latin & \ms{ro fecisse ĩmitatĩom ãtiqͥtatis} \\
        English &  affect the imitation of ancient writers. \\ 
        Archetype & Paradox \\ \midrule
        $P_0$ (Base) & to hear the intonation of the prayers \\
        $P_1$ (OCR) & \textbf{for having imitated the antiquity} \\
        $P_2$ (Corr) & to have advanced the imitation of antiquity \\
        $P_3$ (RAG) & For having conducted an initiation of antiquity \\
        $P_4$ (Full) & to have progressed in the imitation of antiquity \\
    \end{tabular}
\end{table*}

}
{
% \renewcommand{\arraystretch}{1.3}
% \begin{table*}
%     \small\centering
%     \setlength{\tabcolsep}{3pt}\begin{tabular}{m{0.1\linewidth}p{0.8\linewidth}}
%         \textbf{829} & \textbf{Content} \\ \midrule\midrule
%         Image & \includegraphics[width=\linewidth, valign=c]{figures/sample_829.png} \\
%         Latin & \ms{¶ Geometria In numeros atq formas diuisa geometria est quam oratori necessariam esse autumant eruditi: precipit Plato geometriam musicã} \\
%         English &  ¶ Geometry. Geometry is divided into numbers and forms, which the learned assert is necessary for an orator: Plato teaches that geometry and music \\ 
%         Archetype & Cascade \\ \midrule
%         $P_0$ (Base) & Geometry is divided into numerous forms, which is said by scholars to be necessary for an orator. Plato prescribes geometry as \\
%         $P_1$ (OCR) & \textbf{Geometry is divine in numbers and forms, which they declare necessary for the orator. Plato advised studying geometry for music.} \\
%         $P_2$ (Corr) & He boasted that he knew how to manage the republic. The divine promises are duties for trials and for the good, which even the church is. \\
%         $P_3$ (RAG) & Geometry has countless forms. Divine geometry is necessary for the orator, so say the learned. Plato instructs that geometry is closely related to music. \\
%         $P_4$ (Full) & He boasted that he knew how to manage the republic. Commerce with the fearful and bountiful promises are divine, even the church. \\
%     \end{tabular}
% \end{table*}

\begin{table*}
    \small\centering
    \setlength{\tabcolsep}{3pt}\begin{tabular}{m{0.1\linewidth}p{0.8\linewidth}}
        \textbf{667} & \textbf{Content} \\ \midrule\midrule
        Image & \includegraphics[width=\linewidth, valign=c]{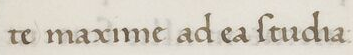} \\
        Latin & \ms{te maxime ad ea studia} \\
        English &  to those studies \\ 
        Archetype & Cascade \\ \midrule
        $P_0$ (Base) & you especially to those pursuits \\
        $P_1$ (OCR) & \textbf{you especially to those studies} \\
        $P_2$ (Corr) & you especially have zeal for these things. \\
        $P_3$ (RAG) & you especially towards those studies \\
        $P_4$ (Full) & you especially towards these pursuits. \\
    \end{tabular}
\end{table*}

% \begin{table*}
%     \small\centering
%     \setlength{\tabcolsep}{3pt}\begin{tabular}{m{0.1\linewidth}p{0.8\linewidth}}
%         \textbf{917} & \textbf{Content} \\ \midrule\midrule
%         Image & \includegraphics[width=\linewidth, valign=c]{figures/sample_917.png} \\
%         Latin & \ms{multis in domo polemarchi. \& in phedro sub procerissima platano considens dixerit. \& in pluribus quoq aliis dialogis.} \\
%         English &  many in the house of Polemarchus. And in the Phaedrus, sitting under a very tall plane tree, he spoke. And in many other dialogues also. \\ 
%         Archetype & Cascade \\ \midrule
%         $P_0$ (Base) & many in the house of the polemarch. But Phaedrus, while trusting in a very tall plane tree, was silent. And in many other dialogues too. \\
%         $P_1$ (OCR) & \textbf{many in the house of Polemarchus and in the portico under a very tall plane tree. And he conducted many dialogues, relying on Phaedrus.} \\
%         $P_2$ (Corr) & to many in the house of Polemarchus and in the presbytery under the most excellent \\
%         $P_3$ (RAG) & Many in the house of Polemarchus and in the sanctuary placed their trust under the very tall plane tree. \\
%         $P_4$ (Full) & In the house of Polemarchus, and in the presbytery he said under the very high plane tree. \\
%     \end{tabular}
% \end{table*}

\begin{table*}
    \small\centering
    \setlength{\tabcolsep}{3pt}\begin{tabular}{m{0.1\linewidth}p{0.8\linewidth}}
        \textbf{572} & \textbf{Content} \\ \midrule\midrule
        Image & \includegraphics[width=\linewidth, valign=c]{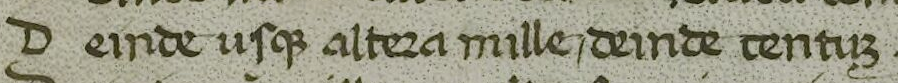} \\
        Latin & \ms{Deinde usq altera mille / deinde centu.} \\
        English &  then another thousand without resting, then a hundred. \\ 
        Archetype & Cascade \\ \midrule
        $P_0$ (Base) & From then on, another thousand are added. \\
        $P_1$ (OCR) & \textbf{Then up to another thousand, then a hundred.} \\
        $P_2$ (Corr) & then up to another thousand then then then then then then then \\
        $P_3$ (RAG) & Then straight on to another thousand, then continuously. \\
        $P_4$ (Full) & Then onwards another thousand, then completed. \\
    \end{tabular}
\end{table*}

\begin{table*}
    \small\centering
    \setlength{\tabcolsep}{3pt}\begin{tabular}{m{0.1\linewidth}p{0.8\linewidth}}
        \textbf{850} & \textbf{Content} \\ \midrule\midrule
        Image & \includegraphics[width=\linewidth, valign=c]{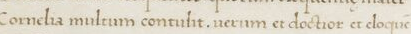} \\
        Latin & \ms{Cornelia multum contulit. uerum et doctior et eloquẽ} \\
        English &  Cornelia contributed much. But Caius was both more learned and more elo- \\ 
        Archetype & Cascade \\ \midrule
        $P_0$ (Base) & Cornelia of many conferred, wise, learned and eloquent. \\
        $P_1$ (OCR) & \textbf{Cornelia contributed much indeed, being both more learned and eloquent.} \\
        $P_2$ (Corr) & Martino and Quintia's misfortunes. Cornelia contributed much, indeed both more learned and eloquent. \\
        $P_3$ (RAG) & Cornelia contributed much, but she was more learned and eloquent. \\
        $P_4$ (Full) & Martino and Quentia of evil. Cornelia contributed much, indeed more learned and eloquent. \\
    \end{tabular}
\end{table*}

\begin{table*}
    \small\centering
    \setlength{\tabcolsep}{3pt}\begin{tabular}{m{0.1\linewidth}p{0.8\linewidth}}
        \textbf{729} & \textbf{Content} \\ \midrule\midrule
        Image & \includegraphics[width=\linewidth, valign=c]{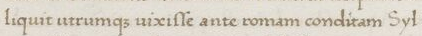} \\
        Latin & \ms{liquit utrumq uixisse ante romam conditam Syl} \\
        English &  that both lived before the founding of Rome, \\ 
        Archetype & Cascade \\ \midrule
        $P_0$ (Base) & he left either to live before the city of Rome, Syl- \\
        $P_1$ (OCR) & \textbf{that each of them lived before the founding of Rome.} \\
        $P_2$ (Corr) & something or both lived before Rome was founded, Syl. \\
        $P_3$ (RAG) & Someone or something, both of them lived before Rome was founded. \\
        $P_4$ (Full) & Something, some others, both lived before Rome was founded, Syl. \\
    \end{tabular}
\end{table*}

\begin{table*}
    \small\centering
    \setlength{\tabcolsep}{3pt}\begin{tabular}{m{0.1\linewidth}p{0.8\linewidth}}
        \textbf{190} & \textbf{Content} \\ \midrule\midrule
        Image & \includegraphics[width=\linewidth, valign=c]{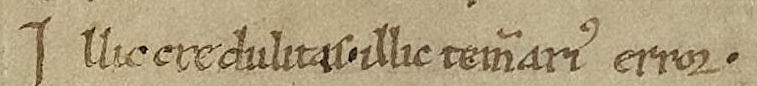} \\
        Latin & \ms{Illic credulitas. illic tem̾ariꝰ error.} \\
        English &  Credulity is there and rash Mistake, \\ 
        Archetype & Cascade \\ \midrule
        $P_0$ (Base) & Where gullibility is, the error follows. \\
        $P_1$ (OCR) & \textbf{There lies credulity, there reckless error.} \\
        $P_2$ (Corr) & There is sweetness there; there is rash error there. \\
        $P_3$ (RAG) & There, credulity; here, rash error. \\
        $P_4$ (Full) & There is credulity; there is a rash error. \\
    \end{tabular}
\end{table*}

\end{document}